\documentclass{article}

\usepackage{microtype}
\usepackage{graphicx}
\usepackage{subcaption}
\usepackage{booktabs} % for professional tables

\usepackage{booktabs}
\usepackage{colortbl}
\usepackage{graphicx}
\usepackage{arydshln}  % for \cdashline
\usepackage{pifont}  % for \ding{55} (✗)
\usepackage{amssymb} % for \checkmark (✓)

\usepackage{comment}
\usepackage{verbatim}
\usepackage{bbm}
\usepackage{amssymb}

\usepackage{xcolor}
\usepackage[most]{tcolorbox}
\usepackage{soul} % \hl 可跨行高亮
\definecolor{hlY}{RGB}{255,255,153}   % 淡黄
\definecolor{hlG}{RGB}{198,239,206}   % 淡绿
\usepackage{mdframed}
\usepackage{xcolor} % 你用到 gray!10 

\usepackage{tikz}
\usetikzlibrary{shapes, arrows, positioning}

\usepackage{hyperref}

\newcommand{\model}{\textsc{ModelTool}}

\usepackage[preprint]{icml2026}
\makeatletter
\renewcommand{\ICML@preprint}{}
\renewcommand{\Notice@String}{}
\renewcommand{\printAffiliationsAndNotice}[1]{\global\icml@noticeprintedtrue%
  \stepcounter{@affiliationcounter}%
  {\let\thefootnote\relax\footnotetext{\hspace*{-\footnotesep}\ificmlshowauthors #1\fi%
      \forloop{@affilnum}{1}{\value{@affilnum} < \value{@affiliationcounter}}{
        \textsuperscript{\arabic{@affilnum}}\ifcsname @affilname\the@affilnum\endcsname%
          \csname @affilname\the@affilnum\endcsname%
        \else
          {\bf AUTHORERR: Missing \textbackslash{}icmlaffiliation.}
        \fi
      }.%
      \ifdefined\icmlcorrespondingauthor@text
         { }Correspondence to: \icmlcorrespondingauthor@text.
      \else
        {\bf AUTHORERR: Missing \textbackslash{}icmlcorrespondingauthor.}
      \fi
    }
  }
}
\makeatother

\usepackage{amsmath}
\usepackage{amssymb}
\usepackage{mathtools}
\usepackage{amsthm}

\usepackage{pifont}
\usepackage{xcolor}
\usepackage{soul}
\definecolor{hlY}{RGB}{255,255,150}  % 黄色
\definecolor{hlG}{RGB}{150,255,150}  % 绿色
\definecolor{hlR}{RGB}{255,180,180}  % 红色

\usepackage[capitalize,noabbrev]{cleveref}

\theoremstyle{plain}

\theoremstyle{definition}

\theoremstyle{remark}

\usepackage[textsize=tiny]{todonotes}

\icmltitlerunning{Robust and Efficient Tool Orchestration via Layered Execution Structures}

\begin{document}

\twocolumn[
  %\icmltitle{Submission and Formatting Instructions for \\
  %  International Conference on Machine Learning (ICML 2026)}
  % \icmltitle{Enabling Tool-Using Agents without Fine-tuning: \\ 
  %   A Dependency-Guided Approach}

      \icmltitle{Robust and Efficient Tool Orchestration via  Layered \\ Execution Structures  with Reflective Correction}

  % It is OKAY to include author information, even for blind submissions: the
  % style file will automatically remove it for you unless you've provided
  % the [accepted] option to the icml2026 package.

  % List of affiliations: The first argument should be a (short) identifier you
  % will use later to specify author affiliations Academic affiliations
  % should list Department, University, City, Region, Country Industry
  % affiliations should list Company, City, Region, Country

  % You can specify symbols, otherwise they are numbered in order. Ideally, you
  % should not use this facility. Affiliations will be numbered in order of
  % appearance and this is the preferred way.
  \icmlsetsymbol{equal}{*}

    \begin{icmlauthorlist}
      %\icmlauthor{Tao Zhe}{equal,ku}
      \icmlauthor{Tao Zhe}{ku}
      \icmlauthor{Haoyu Wang}{nec}
      \icmlauthor{Bo Luo}{ku}
      \icmlauthor{Min Wu}{astar}
      \icmlauthor{Wei Fan}{auckland}
      \icmlauthor{Xiao Luo}{uwm}
      \icmlauthor{Zijun Yao}{ku}
      \icmlauthor{Haifeng Chen}{nec}
      \icmlauthor{Dongjie Wang}{ku}
    \end{icmlauthorlist}
    
    \icmlaffiliation{ku}{Department of Electrical Engineering and Computer Science, University of Kansas, Lawrence, KS, USA}
    \icmlaffiliation{nec}{NEC Laboratories America, Princeton, NJ, USA}
    \icmlaffiliation{astar}{Institute for Infocomm Research, A*STAR, Singapore}
    \icmlaffiliation{auckland}{School of Computer Science, University of Auckland, Auckland, New Zealand}
    \icmlaffiliation{uwm}{Department of Statistics, University of Wisconsin--Madison, Madison, WI, USA}
    
    %\icmlcorrespondingauthor{Tao Zhe}{你的邮箱@ku.edu}
    \icmlcorrespondingauthor{Dongjie Wang}{wangdongjie@ku.edu}
  % You may provide any keywords that you find helpful for describing your
  % paper; these are used to populate the "keywords" metadata in the PDF but
  % will not be shown in the document
  \icmlkeywords{Machine Learning, ICML}

  \vskip 0.3in
]

% this must go after the closing bracket ] following \twocolumn[ ...

% This command actually creates the footnote in the first column listing the
% affiliations and the copyright notice. The command takes one argument, which
% is text to display at the start of the footnote. The \icmlEqualContribution
% command is standard text for equal contribution. Remove it (just {}) if you
% do not need this facility.

\def\model{RETO}

% Use ONE of the following lines. DO NOT remove the command.
% If you have no special notice, KEEP empty braces:
\printAffiliationsAndNotice{}  % no special notice (required even if empty)
% Or, if applicable, use the standard equal contribution text:
% \printAffiliationsAndNotice{\icmlEqualContribution}

\begin{abstract}
Tool invocation is a core capability of agentic systems, yet failures often arise not from individual tool calls but from how multiple tools are organized and executed together. 
Existing approaches tightly couple tool execution with step-wise language reasoning or explicit planning, leading to brittle behavior and high execution overhead.
To overcome these limitations, we revisit tool invocation from the perspective of tool orchestration. 
Our key insight is that effective orchestration does not require precise dependency graphs or fine-grained planning. Instead, a coarse-grained layer structure suffices to provide global guidance, while execution-time errors can be corrected locally.
%
%Specifically, we model tool orchestration as learning a layered execution structure that captures high-level dependencies among tools. This structure induces a layer-wise execution process, decoupling global organization from local tool invocation through context constraints.
Specifically, we model tool orchestration as learning a layered execution structure that captures high-level tool dependencies, inducing layer-wise execution through context constraints.
To handle execution-time failures, we introduce a schema-aware reflective correction mechanism that detects and repairs errors locally.
This design confines errors to individual tool calls and avoids re-planning entire execution trajectories.
This structured execution paradigm enables a lightweight and reusable orchestration component for agentic systems. 
Experimental results show that our approach achieves robust tool execution while reducing execution complexity and overhead. Code will be made publicly available.
%\footnote{ \url{https://github.com/anonymous-submission2026/RETO_code}}.

%Tool invocation is a core capability of agentic systems, yet failures often arise not from individual tool calls but from how multiple tools are organized and executed together. 
%Existing approaches tightly couple tool execution with step-wise language reasoning or explicit planning, leading to brittle behavior and high execution overhead.
%To overcome these limitations, we revisit tool invocation from the perspective of tool orchestration.
%需要转化表述
%Our key insight is that effective orchestration does not require precise planning or exact dependency modeling.
%Instead, it can be achieved by learning approximate execution structures and correcting errors at execution time.
%这里并不能说是structrual imperfection？ 更多是 reliability?
%Specifically, we model tool orchestration as learning a coarse execution structure that captures high-level dependencies among tools. This structure induces a layered execution process, decoupling global organization from local tool invocation. 
%To handle structural imperfections, we introduce reflective correction mechanisms that detect and repair execution errors locally as tools are invoked. This design confines errors to individual tool calls and avoids re-planning entire execution trajectories.
%这段没问题
%This structured execution paradigm enables a lightweight and reusable orchestration component for agentic systems. 
%Experimental results show that our approach achieves robust tool execution while significantly reducing execution complexity and overhead. Code and data are publicly available\footnote{ \url{https://github.com/orbit-913/RETO_code}}.
\end{abstract}

\vspace{-0.5cm}
\begin{figure}[t]
\centering
\includegraphics[width=\columnwidth]{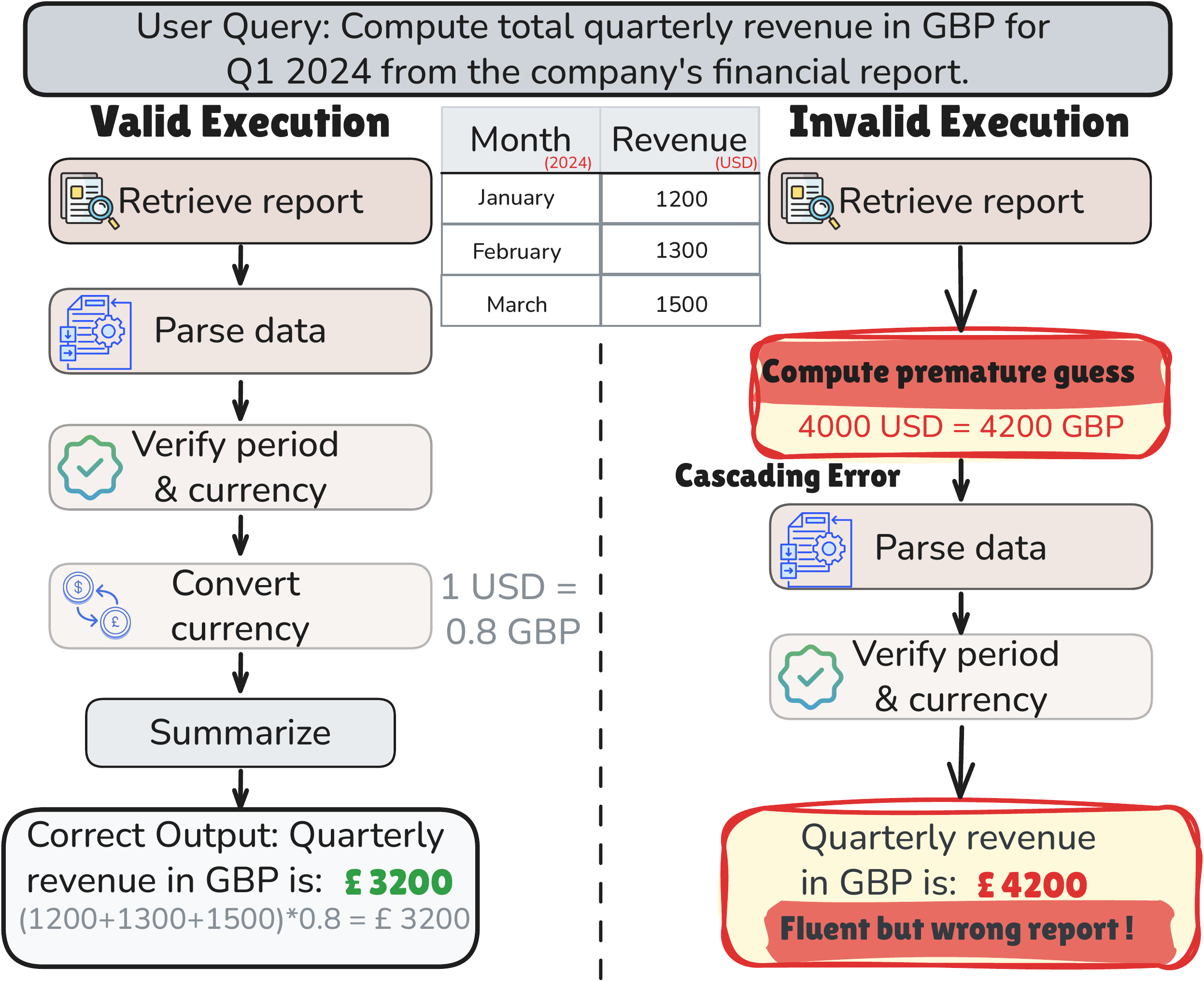}
\caption{Valid vs. invalid tool execution. Hallucinated premature computation from SLM before data parsing leads to cascading errors, producing a fluent but incorrect final answer report.}
\vspace{-0.6cm}
\label{fig:Intro_case}
\end{figure}
\vspace{-0.3cm}
\section{Introduction}
\vspace{-0.2cm}
Large language model (LLM)-based agentic systems have emerged as a dominant paradigm for tackling complex, real-world tasks that require perception, decision-making, and action~\cite{schick2023toolformer,qin2023toolllm}.
At the core of this paradigm is \emph{tool invocation}, which enables agents to interact with external environments and perform operations beyond parametric knowledge.
Recent industry visions, most notably articulated by NVIDIA, point toward hybrid agent architectures that rely on LLMs for high-level planning and small language models (SLMs) for scalable execution~\cite{belcak2025small}.
In this setting, the primary challenge is no longer invoking individual tools, but reliably coordinating multiple tools over long horizons.
As a result, failures increasingly come from \emph{tool orchestration}, rather than individual tool calls.
Errors in ordering and dependency management propagate through execution and lead to fluent but incorrect outcomes, particularly for SLMs with limited context and recovery capacity.~\cite{lu2023chameleon,jiang2025naviagent,mekala2024toolverifier,zhu2025llm}.

For instance, as shown in Figure~\ref{fig:Intro_case}, consider a SLM agent tasked with computing quarterly revenue using multiple tools.
A correct execution verifies data semantics, such as fiscal period and currency, before performing computation, whereas an incorrect execution invokes the calculation tool prematurely on unverified values.
Although the premature computation is numerically valid, it produces a semantically incorrect intermediate result that irreversibly contaminates subsequent execution.
%This example illustrates that tool execution order is a structural constraint that directly determines system verifiability and correctness.
This example illustrates that tool execution order is a structural constraint that directly determines system verifiability and correctness—yet current approaches struggle to enforce such constraints reliably.

%两类传统方法对SLM不好 → fine-tuning方法有效但有成本 → 我们的方法不需要fine-tuning
%Prior tool orchestration approaches largely fall into two paradigms, both of which are suboptimal for small language models (SLMs).
%The first paradigm is \textit{step-wise and reactive} (e.g., ReAct~\cite{yao2022react}, DFSDT~\cite{qin2023toolllm}), where agents decide actions incrementally without an explicit global structure. As a result, these methods often miss implicit prerequisites, leading to dead-end toolchains or excessive computation overhead.
%The second paradigm is \textit{plan--then--execute} (e.g., HuggingGPT~\cite{shen2023hugginggpt}, PS~\cite{wang2023plan}), which constructs a global plan upfront and executes it sequentially. While such plans provide high-level guidance, they are brittle to runtime deviations, such as schema mismatches or empty tool outputs, and typically rely on ad-hoc retries rather than principled repair.
%Consequently, existing approaches lack a unified mechanism that jointly provides \textbf{global structural awareness} and \textbf{local error resilience} during execution.
%%
%Recent work addresses orchestration through SLM fine-tuning. ToolLLaMA with DFSDT~\citep{qin2023toolllm} fine-tunes on tool-use trajectories and employs decision-tree search for multi-step tool invocation. 
%DTA-Tool~\citep{zhu2025divide} constructs DAG-structured supervision from trajectories and fine-tunes LLMs for layer-wise execution. While effective, these approaches incur training costs and couple orchestration capability with SLM parameters.
Prior tool orchestration approaches largely fall into two paradigms, both of which are suboptimal for small language models. 
The first paradigm is \textit{step-wise and reactive} (e.g., ReAct~\citep{yao2022react}), where agents decide actions incrementally 
without an explicit global structure. As a result, these methods often 
miss implicit prerequisites, leading to dead-end toolchains or excessive 
computation overhead. The second paradigm is \textit{plan-then-execute} 
(e.g., HuggingGPT~\citep{shen2023hugginggpt}, PS~\citep{wang2023plan}), 
which constructs a global plan upfront and executes it sequentially. 
While such plans provide high-level guidance, they are brittle to 
runtime deviations, such as schema mismatches, format collapse, or empty tool outputs, 
and typically rely on ad-hoc retries rather than principled repair. 
Recent work addresses these limitations through SLM fine-tuning, 
including decision-tree search~\citep{qin2023toolllm} and layer-wise 
DAG execution~\citep{zhu2025divide}, but incurs training costs. 
Consequently, existing approaches lack a unified mechanism that jointly provides \textbf{global structural awareness} and \textbf{local error resilience} during execution.

To bridge these gaps, our aim is to enable agents to orchestrate multiple tools in a manner that combines global awareness of tool dependencies with robustness to execution-time errors.
Rather than predicting tool calls step by step or following a rigid plan, we ask: \textit{How can structured yet resilient tool execution be achieved, especially without time-consuming fine-tuning of SLMs?}
Achieving this goal, however, involves two key research challenges as follows:

\vspace{-0.3cm}
\begin{itemize}
    \item \textbf{Challenge 1: Fast induction of a coarse execution sketch.} 
    In multi-tool tasks, tool dependencies are implicit and combinatorially complex, making manual specification of execution order prohibitively costly and unscalable. The challenge is to quickly infer a rough execution sketch that captures essential prerequisites without exhaustive planning or tool-specific rules.
    
    \item \textbf{Challenge 2: Faithful execution with effective failure repair.} 
    Even with a reasonable execution sketch, execution-time errors are unavoidable due to noisy tools and hidden constraints.
    The challenge is to identify execution errors and repair them locally under the induced structure, rather than allowing errors to cascade through the chain of tool invocation.
\end{itemize}
\vspace{-0.3cm}

To address these challenges, we propose a \ul{\textbf{R}}obust and \ul{\textbf{E}}fficient \ul{\textbf{T}}ool \ul{\textbf{O}}rchestration framework based on layered execution structures with reflective correction, named \textbf{\model}. 
The key idea is to decouple the global organization from local execution by inducing a coarse execution structure and correcting errors during execution rather than committing to an exact plan.
Specifically, \model\ includes three key stages:
%\emph{Layer-based Tool Orchestration:}
\emph{Learn (layer execution sketch).}
%\model\ infers a rough execution sketch by organizing %tools into a small number of ordered layers based on implicit prerequisite relations from task and tool descriptions.
%长了一点
\model\ employs a lightweight layer predictor to infer a coarse layer structure, organizing tools into ordered layers based on implicit prerequisite relations from task and tool descriptions.
%This sketch provides a sufficient global structure while avoiding the cost and brittleness of precise dependency modeling.
%\emph{Execute (structure-constrained invocation).}
%Guided by the execution sketch, \model\ invokes tools layer by layer, restricting each step to relevant tools and previously validated outputs.
%This constraint reduces decision complexity and limits error propagation, enabling stable execution for small language models.
\emph{Execute (context-constrained invocation).} 
%\emph{Layer-wise Constrained Execution:}
Guided by the layer structure, \model\ invokes tools layer by layer, 
restricting each step to relevant tools and previously validated outputs. 
This constraint reduces decision complexity and limits 
error propagation, enabling stable execution for SLMs.
\emph{Reflect and Repair (local correction).}
%\emph{Reflective Error Recovery:}
%Since execution-time errors are unavoidable, \model\ identifies invalid or inconsistent tool outputs and performs local repair within the existing sketch, rather than replanning globally.
%This design confines failures to individual tool invocations and prevents cascading errors in long multi-tool executions.
Since execution-time errors are unavoidable, \model\ employs schema-aware reflective correction to identify invalid or inconsistent tool outputs and perform local repair within the layer structure, rather than replan globally.
This design confines failures to individual tool calls and prevents cascading errors in long multi-tool executions.
To summarize, our contributions are threefold:

\vspace{-0.3cm}
\begin{itemize}
  \item \textbf{Problem.}
  %We reformulate tool orchestration from step-wise tool selection to structured execution control.
  %This perspective frames orchestration as a system-level execution problem rather than a reasoning task.
  We reformulate tool orchestration from step-wise tool selection to structured execution control via layer structures.
  This perspective frames orchestration as a system-level execution problem.

  % \vspace{-9pt}
  \item \textbf{Algorithm.}
  We propose \model, a robust and efficient tool orchestration framework.
  It combines layered execution structures with reflective local correction to support structured and resilient multi-tool execution.

  % \vspace{-9pt}
  \item \textbf{Evaluation.}
  We evaluate \model\ on large-scale real-world tool benchmarks across different scenarios.
  The results demonstrate consistent improvements in effectiveness, robustness, efficiency, and generalization.

\end{itemize}

\vspace{-0.4cm}
\section{Related Work}
%Related work写作思路
%~\cite{kamoi2024can} detect and self-correct is hard for LLM itself.
\vspace{-0.1cm}
\paragraph{Tool Learning for LLMs.}
Teaching LLMs to use external tools has emerged as a key paradigm for extending model capabilities beyond static parametric knowledge~\cite{schick2023toolformer,qin2023toolllm}.
Early work focused on constructing benchmarks and datasets for tool use~\cite{patil2024gorilla,tang2023toolalpaca,huang2023metatool}.
ToolBench~\cite{qin2023toolllm} introduced a large-scale benchmark with over 16,000 real-world APIs and proposed DFSDT (Depth-First Search-based Decision Tree) for multi-step tool invocation.
Subsequent work improved efficiency through A* search pruning~\cite{zhuang2023toolchain} and compiler-based parallel execution~\cite{kim2024llm}, or enhanced reliability via self-reflection~\cite{du2024anytool} and preference optimization~\cite{chen2024advancing}. However, hallucination and instability still hinder reliable tool invocation~\cite{patil2025bfcl}.
%加一点
%\vspace{-8pt}
\vspace{-0.5cm}
\paragraph{Planning Paradigms for Tool Use.}
%Existing approaches to tool orchestration fall into two categories.
Prior work on tool orchestration can be broadly grouped into two paradigms by how it organizes tool dependencies during execution.
\textit{Reactive methods} like ReAct~\cite{yao2022react} and Reflexion~\cite{shinn2023reflexion} interleave reasoning and action step-by-step, offering flexibility but lacking global dependency awareness.
\textit{Plan-then-execute methods} like HuggingGPT~\cite{shen2023hugginggpt} and Plan-and-Solve~\cite{wang2023plan,xu2023rewoo} first generate a complete plan before execution, but struggle with runtime failures due to the disconnect between planning and execution. Recent work~\cite{wu2024can} reveals that attention biases and auto-regressive loss fundamentally limit LLMs' ability to navigate graph-structured decisions. 
%and proposes integrating graph neural networks to assist planning.
%Our work bridges these paradigms: we predict a coarse-grained dependency structure (layer assignments) that provides global awareness while allowing flexible argument generation at execution time.
Our work bridges these paradigms: we offload coarse-grained layer prediction to a lightweight module, providing global dependency awareness while keeping execution locally adaptive to runtime failures.

\vspace{-15pt}
\paragraph{Efficient Tool Learning with Smaller Models.}
Smaller language models offer computational efficiency but struggle with multi-tool orchestration due to limited context capacity and weaker long-horizon reasoning~\cite{belcak2025small}.
Prior work addresses this by internalizing tool-related knowledge into LLMs through fine-tuning. 
DTA-Tool~\cite{zhu2025divide} transforms sequential trajectories into DAG structures and fine-tunes LLMs for parallel execution. 
GAP~\cite{wu2025gap} combines supervised fine-tuning with reinforcement learning to generate dependency graphs.
ToolGen~\citep{wang2024toolgen} 
fine-tunes LLMs to internalize tool knowledge for autonomous retrieval and execution.
Shen et al.~\citep{shen2024small} distribute tasks across 
multiple fine-tuned agents. While effective, these approaches rely on LLM fine-tuning, incurring training cost and limiting adaptability to evolving tool sets.
Our work takes a different path: we \textit{offload planning to a lightweight external module} and use \textit{context-constrained execution} to reduce the per-step reasoning burden on off-the-shelf SLMs, enabling competitive performance \textit{without any LLM fine-tuning}.
\vspace{-2pt}

%Method Section（对应 P6）
%3.1 Rough层级结构如何预测（模型、输入表示、输出形式）
%3.2 Layer-wise execution（如何推进、同层并行与否、输出验证器）
%3.3 Reflect/repair loop（触发条件、诊断模板、修复动作、停止条件）
\vspace{-2pt}
\section{Methodology}

\begin{figure*}[t]
    \centering
    \includegraphics[width=0.95\linewidth]{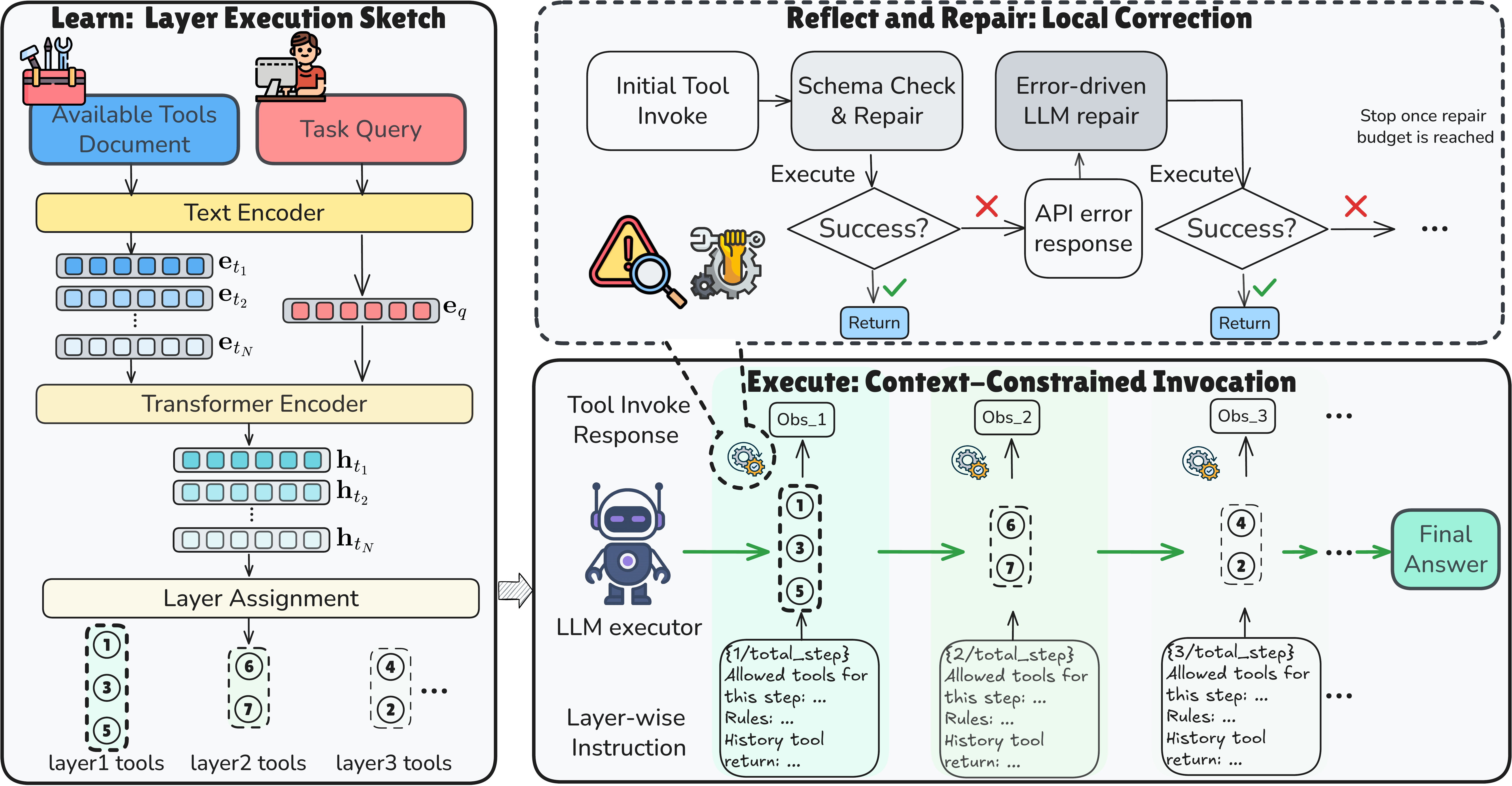}
    \caption{\textbf{Overview of \model.} Left: layer assignment from the task query and tool descriptions. Top-right: repair loop for failed tool calls (schema gate $\rightarrow$ LLM repair) under a budget. Bottom-right: context-constrained execution with step-specific tool constraints; observations are carried across layers to produce the final answer.}
    \label{fig:framework}
\end{figure*}

\vspace{-2pt}
\subsection{Framework Overview}
\vspace{-2pt}
% Standard step-wise paradigms such as ReAct ask a single model to plan over long horizons, select tools from a large inventory, and generate precise arguments in one coupled loop.
% This works reasonably well for large proprietary models.
% Small language models (SLMs), however, struggle under this load: they lose global task coherence in long traces and tool descriptions, even though they remain reliable on local, atomic tool calls when the context is short.
Figure~\ref{fig:framework} gives an overview of \model.
At a high level, \model\ separates global structure from local execution for small LM agents.
It consists of three components:
(1) \textbf{Learn: Layer Execution Sketch}, which predicts a coarse execution layer for each tool, providing a global scaffold without enumerating all pairwise edges;
(2) \textbf{Execute: Context-Constrained Invocation}, which runs tools layer by layer and restricts the SLM to tools in the current layer, keeping the effective context short; and
(3) \textbf{Reflect and Repair: Local Correction}, which monitors tool outputs with error checks and triggers local correction when a call fails or returns uninformative results.
This design moves global planning into a lightweight structural predictor, while the SLM focuses on short-horizon tool use with explicit failure handling.

%\subsection{Layer-based Tool Orchestration}
\vspace{-5pt}
\subsection{Learn: Layer Execution Sketch}
\label{sec:layer_orchestration}
\vspace{-1.5 pt}
%问题 + 失败机理（failure mechanism）
%Multi-tool tasks often fail when the model commits to premature steps before verifying prerequisite information; such ordering errors can then cascade through subsequent calls.
Multi-tool tasks often fail when the model violates prerequisite dependencies, causing downstream errors.
%为什么不做 full plan/DAG：显式依赖图/完整计划在大工具集下成本高、且对早期错误敏感（brittle）
%Explicitly constructing a full dependency graph or a complete plan can be expensive and brittle in practice. It requires many fine-grained dependency decisions.
Explicitly constructing a full dependency graph or complete plan is expensive and brittle in practice, requiring many fine-grained decisions.
%我们的替代物是什么：只学一个粗粒度的执行骨架——给每个 tool 预测一个 layer。
Instead, we assign each tool to an execution layer to form a coarse execution sketch. A tool in layer $k$ may depend on outputs from layers $<k$.
%Tools in the same layer have no required precedence.
%为什么这样足够且不会 overclaim：layer 只编码必要的 precedence；细粒度数据流依赖留到 runtime 用 schema/I/O 去对齐。
During execution, the model fills tool arguments based on tool schemas and outputs from earlier layers.
This enables fine-grained information flow without predicting pairwise dependency edges. Tools within a layer can be executed in parallel.
%Settings
Formally, given a user query $q$ and candidate tools $\mathcal{T}=\{t_1,\dots,t_N\}$, 
we predict an execution layer $l_i\in\{0,\dots,L-1\}$ for each tool $t_i$, 
where $L$ is the maximum number of layers. 
$l_i=0$ means $t_i$ is executed first in the coarse order, 
and $l_i=k$ means $t_i$ may depend on outputs from layers $<k$.
%Settings(what)
%Given a user query $q$ and a set of candidate tools $\mathcal{T} = \{t_1, t_2, \dots, t_N\}$, our goal is to predict an execution layer $l_i \in \{0, 1, \dots, L - 1\}$ for each tool $t_i$, where $L$ denotes the maximum number of layers. A layer assignment $l_i = 0$ indicates that tool $t_i$ should be executed first (i.e., it has no dependencies), while $l_i = k$ indicates that $t_i$ depends on outputs from tools in layers $< k$. 
%Motivation
%This layer-based formulation encodes a valid topological order: tools assigned to layer $k$ may depend on some tools in earlier layers, while tools within the same layer are mutually independent and thus parallelizable. The exact data dependencies are resolved at runtime based on actual input/output matching. 

\vspace{-5pt}
\paragraph{Input Representation.}
We embed the query $q$ and each tool document $\mathrm{doc}(t_i)$ into a shared $d$-dimensional space using a pre-trained text encoder:
%\vspace{-3pt}
\begin{align}
\mathbf{e}_q &= \mathrm{Encoder}(q) \in \mathbb{R}^d, \\
\mathbf{e}_{t_i} &= \mathrm{Encoder}(\mathrm{doc}(t_i)) \in \mathbb{R}^d,
\end{align}
where $\mathrm{doc}(t_i)$ is a textualized tool document, formed by concatenating the tool name, description, and parameter schema (and optionally other metadata when available).

\vspace{-3pt}
\paragraph{Layer Predictor.}
We design a lightweight neural network that takes 
query and tool embeddings as input and predicts layer assignments. The model consists of three stages:
\textit{\mbox{(1) Projection.}} 
Since the query and tools play different semantic roles, we apply separate projections with LayerNorm and ReLU:
\begin{align}
\mathbf{z}_q &= \mathrm{ReLU}(\mathrm{LN}(\mathbf{W}_q \mathbf{e}_q)) \in \mathbb{R}^{d'}, \\
\mathbf{z}_{t_i} &= \mathrm{ReLU}(\mathrm{LN}(\mathbf{W}_t \mathbf{e}_{t_i})) \in \mathbb{R}^{d'} .
\end{align}
where $\mathbf{W}_q,\mathbf{W}_t \in \mathbb{R}^{d'\times d}$ are learnable and $\mathrm{LN}$ denotes LayerNorm.
We concatenate them into a sequence
\begin{equation}
\mathbf{X} = [\mathbf{z}_q;\mathbf{z}_{t_1};\dots;\mathbf{z}_{t_N}] \in \mathbb{R}^{(1+N)\times d'} .
\end{equation}
\textit{\mbox{(2) Contextual Encoding.}} The concatenated sequence is processed 
by a Transformer Encoder, allowing the query and tools to exchange information 
via self-attention:
\begin{align}
    \mathbf{H} &= \mathrm{TransformerEncoder}(\mathbf{X}) \in \mathbb{R}^{(1+N) \times d'} \\
    \mathbf{h}_{t_i} &= \mathbf{H}[i+1] \in \mathbb{R}^{d'}, \quad i = 1, \dots, N
\end{align}
Through this mechanism, each tool representation $\mathbf{h}_{t_i}$\ integrates 
task semantics from the query and functional information from other candidate tools, enabling the model to capture dependency cues among tools under the query.

\textit{\mbox{(3) Ordinal Output.}} We cast layer prediction as ordinal regression~\cite{cao2020rank}
to exploit the inherent layer order. For $L$ layers, we predict $L{-}1$
cumulative probabilities:
\begin{equation}
    P(l_i > k) = \sigma(\mathbf{w}^{\top}\mathbf{h}_{t_i} + b_k), \quad k=0,\dots,L-2 ,
\end{equation}
where $\sigma$ is the sigmoid, $\mathbf{w}\in\mathbb{R}^{d'}$ is shared across thresholds,
and $b_k$ are learnable biases. We decode the layer by
\begin{equation}
    \hat{l}_i = \sum_{k=0}^{L-2} \mathbbm{1}\!\left[P(l_i>k) > 0.5\right].
    \label{eq:decode}
\end{equation}

\paragraph{Training.} Given ground-truth layer $l_i^*$, we construct cumulative 
binary labels $y_i^{(k)} = \mathbbm{1}[l_i^* > k]$ and minimize weighted 
binary cross-entropy:
\begin{equation}
    \mathcal{L} = \frac{1}{N} \sum_{i=1}^{N} \sum_{k=0}^{L-2} 
    \lambda_k \cdot \text{BCE}(P(l_i > k),\, y_i^{(k)})  
\end{equation}
where $\lambda_k$ balances class imbalance at each threshold.

\paragraph{Inference.} 
The predicted layer follows Eq.~\eqref{eq:decode}, counting exceeded thresholds. The predictor additionally enforces schema-level constraints: if a tool's required input matches another tool's output type in the available document, it is shifted to a subsequent layer to ensure valid execution order.

%\subsection{Layer-wise Constrained Execution}
\subsection{Execute: Context-Constrained Invocation}
% === 开头：motivation + 核心 idea ===
%列得要“像事实描述/设计动机”，而不是像在宣称你解决了三大难题
Given the layer assignments from \S\ref{sec:layer_orchestration}, we execute tools one layer at a time.
At layer $k$, the LM only sees the schemas of the current-layer tools and the tool outputs from earlier layers, along with brief execution instructions.
This design prevents out-of-turn tool calls, keeps the effective context compact, and breaks multi-step reasoning into per-layer decisions. It is particularly helpful for small models in long-context settings~\cite{liu2024lost,levy2024same}.
\vspace{-2pt}
\paragraph{Execution Protocol.}
Let $\mathcal{T}_k = \{t_i \mid l_i = k\}$ denote the set of tools assigned to layer $k$. For each layer $k = 0,1,\ldots,L-1$, we construct a prompt that includes the current layer index $k$ (out of $L$), the schemas of $\mathcal{T}_k$, the original query, prior tool outputs, and brief execution instructions.
The LM generates tool calls with arguments; we execute same-layer calls in parallel under the layer plan and append observations for subsequent layers.
We enforce the constraint by omitting tools outside $\mathcal{T}_k$ from the tool list, preventing out-of-turn calls by construction.
The LM retains autonomy over \emph{how many times} to call each tool and with what arguments.
The layer assignment only determines \emph{when} a tool becomes available to the LM agent.
% ===最终答案 ===
\vspace{-5pt}
\paragraph{Answer Synthesis.}
Finally, after all layers execution is complete, we force a dedicated \texttt{Finish} call.
The LLM receives an execution summary of all observations and must 
produce a \texttt{final\_answer} grounded only in this summary. We 
explicitly instruct: ``if required information is missing, acknowledge 
the limitation.'' This prevents hallucinating results for tools that 
failed or were never executed.

\vspace{-1pt}
\subsection{Reflect and Repair: Local Correction}
\vspace{-1pt}
Even with a correct tool plan, small LMs often produce invalid tool calls: arguments may miss required fields, contain type or enum errors, or trigger runtime failures under hidden constraints (e.g., unexpected keyword arguments, validation errors).
If such calls are executed as-is, they pollute the context with noisy error messages and make later reasoning depend on corrupted observations.
To both prevent this pollution and salvage useful calls when possible, we wrap each tool call in a \emph{reflect--then--repair} mechanism that (i) validates arguments against tool schemas before execution and (ii) performs error-driven local repair under a fixed budget.
This localized reflection avoids costly trajectory-level retries while preventing error accumulation across layers.

\paragraph{Schema Gate.}
For each tool $t$, we build a JSON-schema index $S_t$ from its ToolBench
specification (required keys, top-level properties, primitive types, and
enums).
Given a tool call with arguments $A$, the Schema Gate applies a lightweight
validator:
\begin{equation}
    \text{Gate}(A, S_t) =
    \begin{cases}
    \text{ACCEPT}, & \text{if } \text{Valid}(A, S_t) \\
    \text{REJECT}, & \text{otherwise},
    \end{cases}
\end{equation}

where $\text{Valid}$ checks that all required keys are present and that the
types and enum values of existing keys match the schema.
Unknown top-level keys are simply dropped.
Accepted calls are executed normally; rejected calls are skipped and passed
to the repair module together with a structured error report
(missing fields, type errors, enum violations, and dropped keys).
This gate removes many systematic argument errors before they reach the
environment, while remaining cheap and tool-agnostic.

\paragraph{Repair mechanism.}
As shown in algorithm~\ref{alg:repair}. For each rejected or failed call $(t, A)$ with schema $S_t$ and diagnosis $\mathcal{D}$, we apply an error-driven repair operator $R$ under a per-trajectory budget $B$. $R$ first tries cheap deterministic edits (e.g., removing unexpected keys). If these fail and the diagnosis suggests argument issues, it falls back to an LLM-driven repair based on the error message. All repaired calls are re-validated by the Schema Gate before execution. Once $B$ is exhausted, errors are logged but not retried.
%Compared to naive retries, this reflect--then--repair operator isolates errors at the level of individual tool calls rather than entire trajectories. Repairs are applied locally and kept off the main dialogue trace, so transient argument or tool errors do not bloat the context or bias subsequent reasoning. This prevents malformed arguments from propagating across layers and provides a controlled way to use the LM for targeted correction instead of re-planning the whole toolchain.
%Compared to naive retries, this operator isolates errors at individual tool calls: repairs are kept off the main dialogue trace, preventing context pollution and avoiding trajectory-level re-planning.
Compared to naive retries, this reflect–then–repair operator isolates errors at individual tool calls rather than entire trajectories. Repairs are kept off the main dialogue trace, preventing context pollution and avoiding global re-planning.

\begin{algorithm}[t]
\caption{Error-driven repair operator $R$}
\label{alg:repair}
\begin{algorithmic}[1]
\REQUIRE Tool $t$, arguments $A$, schema $S_t$, diagnosis $\mathcal{D}$, budget $B$
\ENSURE Repaired call $(t, A')$ or $\bot$
\IF{$B = 0$}
    \STATE \textbf{return} $\bot$
\ENDIF
\STATE \textit{\# Tier 1: deterministic repair (no LLM cost)}
\STATE $A' \gets R_{\mathrm{det}}(A, \mathcal{D})$ \COMMENT{e.g., drop unexpected keys}
\IF{$A' \neq \bot$ \AND $\text{Gate}(A', S_t) = \textsc{Accept}$}
    \STATE \textbf{return} $(t, A')$
\ENDIF
\STATE \textit{\# Tier 2: LLM-driven repair}
\STATE $A' \gets R_{\mathrm{LLM}}(A, S_t, \mathcal{D})$
\IF{$A' \neq \bot$ \AND $\text{Gate}(A', S_t) = \textsc{Accept}$}
    \STATE \textbf{return} $(t, A')$
\ENDIF
\STATE \textbf{return} $\bot$
\end{algorithmic}
\end{algorithm}
\vspace{-6pt}

\vspace{-5pt}
\section{Experiment}
\vspace{-1pt}
\subsection{Experiment Settings}
%\paragraph{Datasets.}
\textbf{Datasets.}
We evaluate on StableToolBench~\cite{guo2024stabletoolbench}, which provides a caching system and API simulator to mitigate instability of real-time APIs while preserving the task distribution of ToolBench~\cite{qin2023toolllm}. Test cases span three generalization dimensions: \textit{Inst.}~(unseen instructions, same tools), \textit{Tool}~(unseen tools, same category), and \textit{Cat.}~(unseen tools, unseen category). Task complexity varies across three scenarios: \textit{I1}~(single-tool), \textit{I2}~(intra-category multi-tool), and \textit{I3}~(intra-collection multi-tool), with difficulty increasing from I1-Inst.\ to I3-Inst.
To train the layer predictor, we derive supervision from the training portion of ToolBench. We reuse the structured execution trajectories released with DTA-Tool~\cite{zhu2025divide}, mapping each tool call to a layer index based on its structure in the execution tree, and then split the resulting layer-labeled data into training, validation, and test sets for model selection.
\vspace{-5pt}
\paragraph{Baselines and Models.}
We compare three execution strategies: ReAct~\cite{yao2022react}, 
DFSDT~\cite{qin2023toolllm}, and our plan-then-execute approach. 
To evaluate across model scales and training regimes, we test on: 
(1)~\textit{non-tool-tuned models}: Qwen2.5-7B-Instruct~\cite{qwen25_report} 
and LLaMA-3.1-8B-Instruct~\cite{meta_llama31_blog}; 
(2)~\textit{tool-tuned model}: ToolLLaMA~\cite{qin2023toolllm},  which is fine-tuned on the ToolBench training set for tool use; and 
(3)~\textit{closed-source models}: GPT-3.5-0613 and GPT-3.5-1106, as a reference for proprietary model performance. 
%To further examine how the SLM scale affects performance under our framework, we also evaluate Qwen2.5-3B-Instruct, Qwen2.5-1.5B-Instruct, and Qwen2.5-0.5B-Instruct.
To further examine how SLM scale affects performance, we also evaluate Qwen2.5-3B/1.5B/0.5B-Instruct.
\vspace{-1pt}
%\paragraph{Evaluation Metrics}

\begin{table*}[t]
\centering
\caption{Solvable Pass Rate (\%) on StableToolBench. Best results among open-source models are in \textbf{bold}; \underline{underline} denotes the second best. \textit{Tool-tuned} indicates whether the backbone is tool-tuned (\checkmark) or untuned (\ding{55}) ; -- denotes closed-source baselines.}
%\vspace{-1pt}
\label{tab:main_results}
\resizebox{\textwidth}{!}{
\begin{tabular}{llcccccccc}
\toprule
\textbf{Method} & \textbf{Backbone} & \textbf{Tool-tuned} & \textbf{I1-Inst.} & \textbf{I1-Tool} & \textbf{I1-Cat.} & \textbf{I2-Inst.} & \textbf{I2-Cat.} & \textbf{I3-Inst.} & \textbf{Avg.} \\
\midrule
\multicolumn{10}{c}{\cellcolor{gray!25}\textit{Closed-source LLMs}} \\
\noalign{\vskip 0.3em}
ReAct & GPT-3.5-0613 & -- & 50.1\tiny{±0.6} & 51.6\tiny{±0.4} & 49.8\tiny{±0.3} & 37.3\tiny{±0.8} & 40.3\tiny{±0.7} & 43.2\tiny{±1.5} & 45.4 \\
ReAct & GPT-3.5-1106 & -- & 49.8\tiny{±1.2} & 51.3\tiny{±1.3} & 40.1\tiny{±0.3} & 39.8\tiny{±1.2} & 49.2\tiny{±0.7} & 59.6\tiny{±0.8} & 48.3 \\
\midrule
\multicolumn{10}{c}{\cellcolor{gray!25}\textit{Open-source LLMs}} \\
\noalign{\vskip 0.3em}
ReAct & ToolLLaMA-7B & \checkmark & 31.5\tiny{±0.8} & 34.0\tiny{±0.4} & 39.5\tiny{±0.5} & 25.5\tiny{±0.4} & 34.5\tiny{±0.8} & 32.8\tiny{±0.0} & 33.0 \\
\rowcolor{green!10}DFSDT & ToolLLaMA-7B & \checkmark & \underline{48.8\tiny{±2.4}} & \underline{48.9\tiny{±1.0}} & \underline{48.9\tiny{±0.4}} & \underline{37.3\tiny{±0.4}} & \underline{45.8\tiny{±1.0}} & \textbf{53.0\tiny{±2.0}} & \underline{47.1} \\
\noalign{\vskip 0.3em}
\hdashline
\noalign{\vskip 0.3em}
ReAct & LLaMA-3.1-8B & \ding{55} & 0.0\tiny{±0.0} & 0.0\tiny{±0.0} & 0.0\tiny{±0.0} & 0.0\tiny{±0.0} & 0.0\tiny{±0.0} & 0.0\tiny{±0.0} & 0.0 \\
DFSDT & LLaMA-3.1-8B & \ding{55} & 0.0\tiny{±0.0} & 0.0\tiny{±0.0} & 0.0\tiny{±0.0} & 0.0\tiny{±0.0} & 0.0\tiny{±0.0} & 0.0\tiny{±0.0} & 0.0 \\
\model\ (Ours) & LLaMA-3.1-8B & \ding{55} & 14.0\tiny{±0.8} & 17.4\tiny{±1.6} & 22.9\tiny{±0.6} & 16.7\tiny{±0.8} & 16.8\tiny{±1.3} & 14.5\tiny{±1.9} & 17.1 \\
\noalign{\vskip 0.3em}
\hdashline
\noalign{\vskip 0.3em}
ReAct & Qwen2.5-7B & \ding{55} & 8.4\tiny{±0.3} & 11.0\tiny{±0.3} & 8.1\tiny{±0.3} & 11.3\tiny{±0.8} & 17.2\tiny{±0.8} & 1.1\tiny{±0.8} & 9.5 \\
DFSDT & Qwen2.5-7B & \ding{55} & 5.7\tiny{±0.3} & 15.3\tiny{±0.3} & 10.1\tiny{±0.3} & 17.1\tiny{±0.8} & 24.7\tiny{±0.8} & 0.5\tiny{±0.4} & 12.2 \\
\rowcolor{blue!10} \model\ (Ours) & Qwen2.5-7B & \ding{55} & \textbf{50.3\tiny{±0.5}} & \textbf{50.9\tiny{±0.6}} & \textbf{54.4\tiny{±1.0}} & \textbf{46.5\tiny{±1.8}} & \textbf{52.4\tiny{±1.1}} & \underline{42.5\tiny{±0.7}} & \textbf{49.5} \\
\bottomrule
\end{tabular}
}
\vspace{-10pt}
\end{table*}

\textbf{Evaluation Metrics.}
Following the StableToolBench work~\cite{guo2024stabletoolbench}, we report two metrics: (1) Solvable Pass Rate (SoPR): the percentage of 
solvable tasks that the model successfully completes, as judged by 
GPT-4o; and (2) Solvable Win Rate (SoWR): the percentage of 
tasks where the model's output is preferred over a GPT-3.5-0613+ReAct 
baseline, using GPT-4o as the judge. SoWR is computed as 
$(N_{\text{win}} + N_{\text{tie}}/2) / N_{\text{total}}$, where 
$N_{\text{win}}$, $N_{\text{tie}}$, and $N_{\text{total}}$ denote the 
number of wins, ties, and total comparisons, respectively. We average 
results over three runs to mitigate the variance from LLM-based automatic evaluation.

\subsection{Experimental Results}
\vspace{-1pt}
\subsubsection{Overall performance}
\vspace{-1pt}
%RQ1 有效性，是否整体success rate 更高？

%question
This experiment aims to answer: \textit{Whether our method can lift non-tool-tuned SLMs to a Solvable Pass Rate comparable to tool-tuned and proprietary agents.}
%exp setting
To this end, we evaluate three reasoning methods (ReAct, DFSDT, and our \model) across different model backbones, including non-tool-tuned open-source models (Qwen2.5-7B, LLaMA-3.1-8B), a tool-tuned model (ToolLLaMA-7B), and closed-source models (GPT-3.5). We report SoPR evaluated by GPT-4o with three independent runs (Table~\ref{tab:main_results}).
%observation
We observe three findings.
%First, traditional methods fail on non-tool-tuned models: ReAct and DFSDT yield low SoPR on both LLaMA-3.1-8B and Qwen2.5-7B.
First, traditional methods perform poorly on non-tool-tuned models: ReAct and DFSDT yield low SoPR on both backbones.
Second, \model\ improves non-tool-tuned models, lifting 
Qwen2.5-7B from 12.2\% (DFSDT) to 49.5\%, and LLaMA-3.1-8B from 0.0\% to 17.1\%.
Third, \model\ on Qwen2.5-7B surpasses tool-tuned ToolLLaMA-7B and exceeds both GPT-3.5 ReAct baselines.
%analysis
The key driver is that ReAct and DFSDT 
require a single model to handle long-horizon reasoning in one trace, 
where early mistakes cascade through subsequent steps. 
\model\ separates these concerns: a planner provides a coarse 
dependency sketch, and the LM operates within this scaffold with 
schema checks and local repair. This turns entangled reasoning into 
short, well-scoped decisions that match the effective context window 
of non-tool-tuned SLMs.
%conclusion
In summary, \model\ enables non-tool-tuned open-source models to reach performance comparable to tool-tuned counterparts and proprietary LLMs.
%
% \vspace{-10 pt}
\begin{figure}[t]
\centering
\includegraphics[width=\columnwidth]{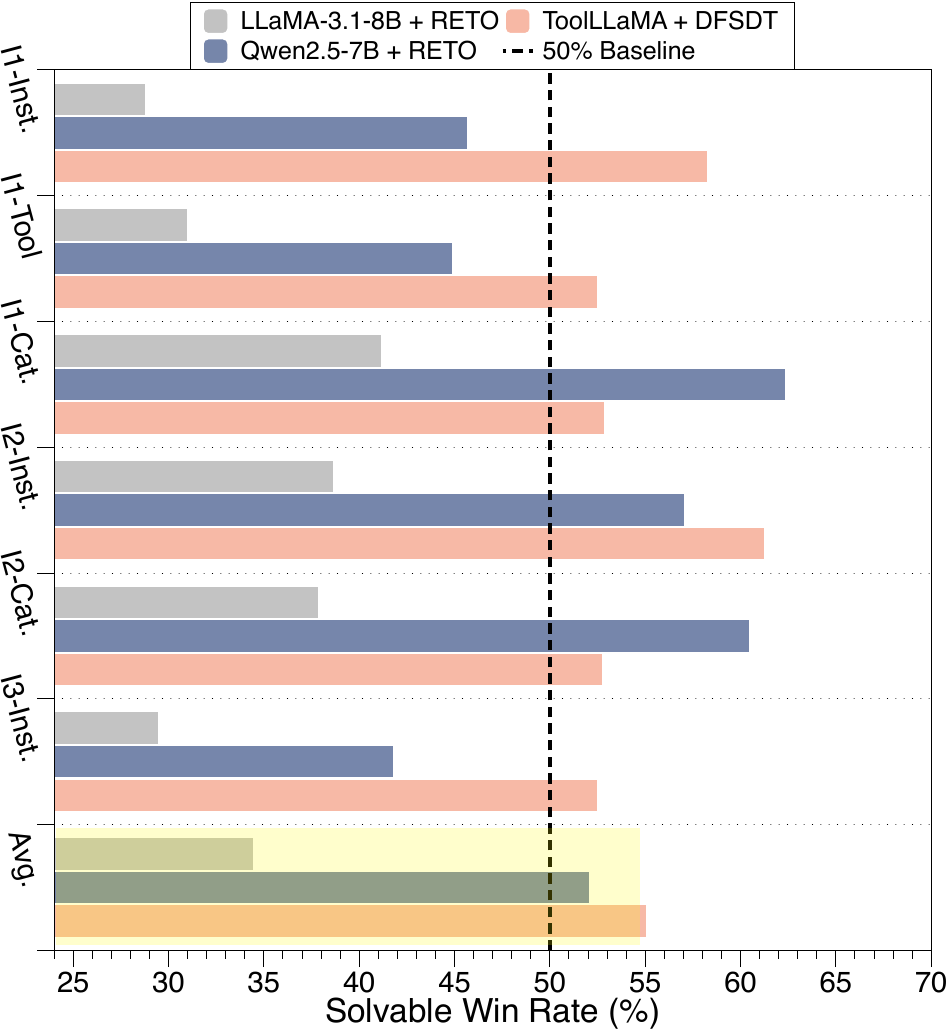}
\caption{Solvable Win Rate (\%) of open-source models against GPT-3.5-0613 + ReAct on StableToolBench, computed from per-instance comparisons; the dashed line marks 50\% parity.}
\label{fig:win_rate}
% \vspace{-15pt}
\end{figure}
\begin{figure*}[t]
  \centering
  \begin{subfigure}[t]{0.24\textwidth}\centering
    \includegraphics[width=\linewidth]{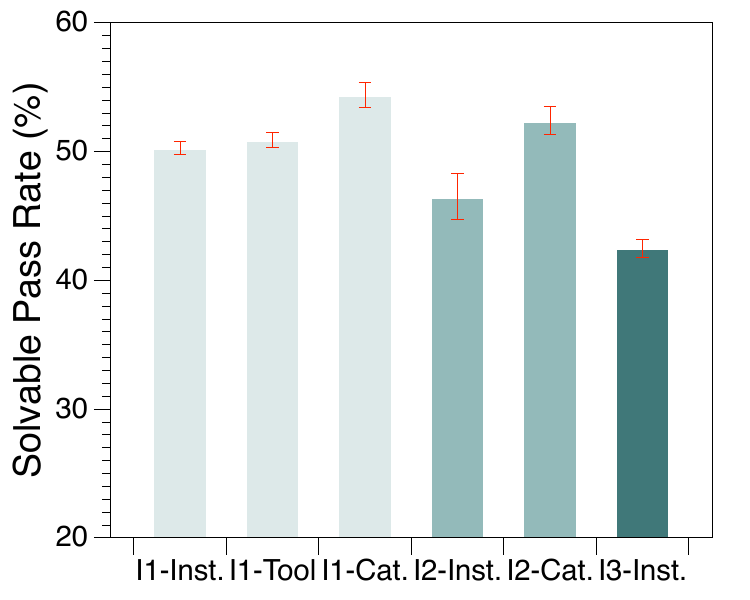}\caption{Qwen2.5-7B}
  \end{subfigure}\hfill
  \begin{subfigure}[t]{0.24\textwidth}\centering
    \includegraphics[width=\linewidth]{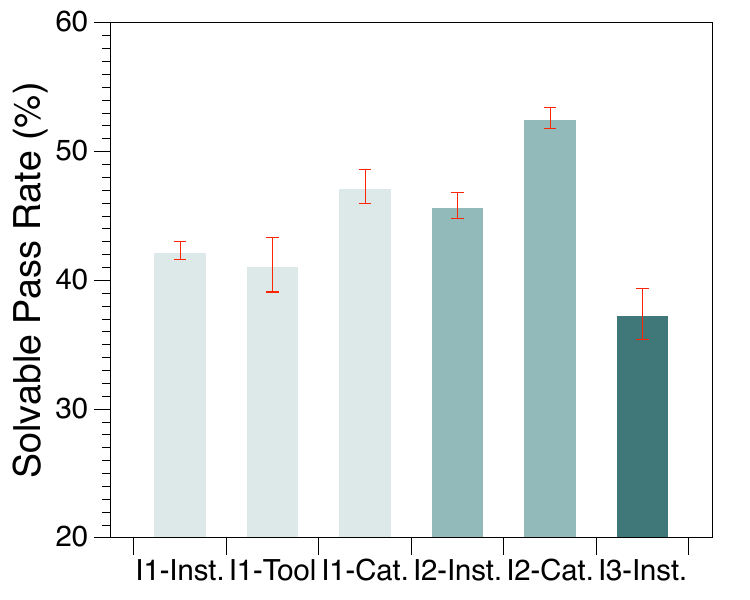}\caption{Qwen2.5-3B}
  \end{subfigure}\hfill
  \begin{subfigure}[t]{0.24\textwidth}\centering
    \includegraphics[width=\linewidth]{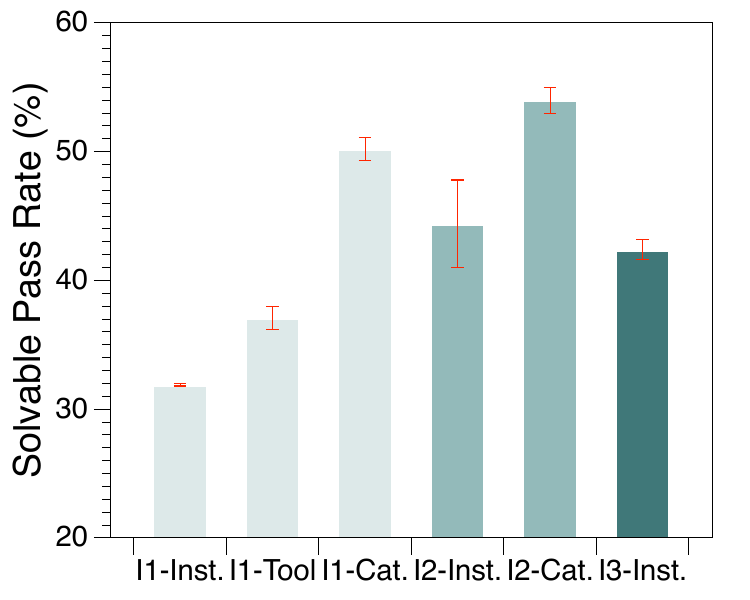}\caption{Qwen2.5-1.5B}
  \end{subfigure}\hfill
  \begin{subfigure}[t]{0.24\textwidth}\centering
    \includegraphics[width=\linewidth]{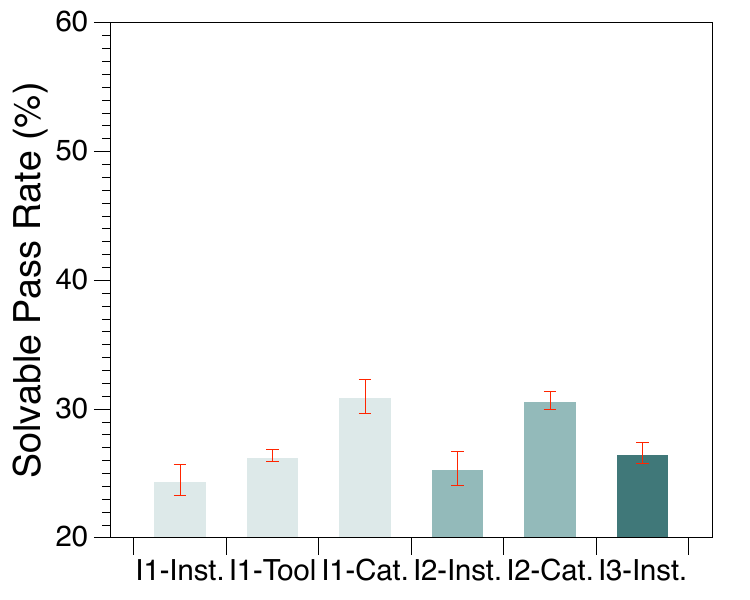}\caption{Qwen2.5-0.5B}
  \end{subfigure}
  \caption{SoPR (\%) across Qwen2.5 model sizes (0.5B–7B). Performance degrades gradually, with 3B and 1.5B remaining functional.}
  \label{fig:size}
\end{figure*}
%\vspace{1pt}
\subsubsection{Answer quality evaluation}
%question
This experiment aims to answer: \textit{whether our method delivers tool-tuned-level answer quality from non-tool-tuned SLMs.}
%settings
To this end, we follow the StableToolBench protocol to report Solvable Win Rate. For each test instance, GPT-4o compares a model's answer with that of GPT-3.5-0613 + ReAct. We evaluate three configurations: ToolLLaMA-7B + DFSDT (tool-tuned), Qwen2.5-7B + \model\, and LLaMA-3.1-8B + \model.
%observation
As shown in Figure~\ref{fig:win_rate}, non-tool-tuned Qwen2.5-7B with \model\ is comparable to tool-tuned ToolLLaMA-7B + DFSDT in average win rate, with both configurations exceeding 50\% parity on several subsets. LLaMA-3.1-8B, however, lags behind. (For details before merging the tie label, please refer to Appendix~\ref{app:win_lose_tie}.)
%analysis
These results indicate that \model\ helps bridge the gap between non-tool-tuned and tool-tuned SLMs. The performance gap between Qwen2.5-7B and LLaMA-3.1-8B may stem from 
%\vspace{-18pt}
differences in instruction-following ability, as faithful execution of the execution sketch depends on the backbone's capability to follow instructions.
%conclusion
In summary, our method lifts capable non-tool-tuned SLMs to match tool-tuned baselines in answer quality when evaluated against GPT-3.5-0613.
\begin{figure}[t]
  \vspace{-10pt}
  \centering
  \includegraphics[width=\columnwidth]{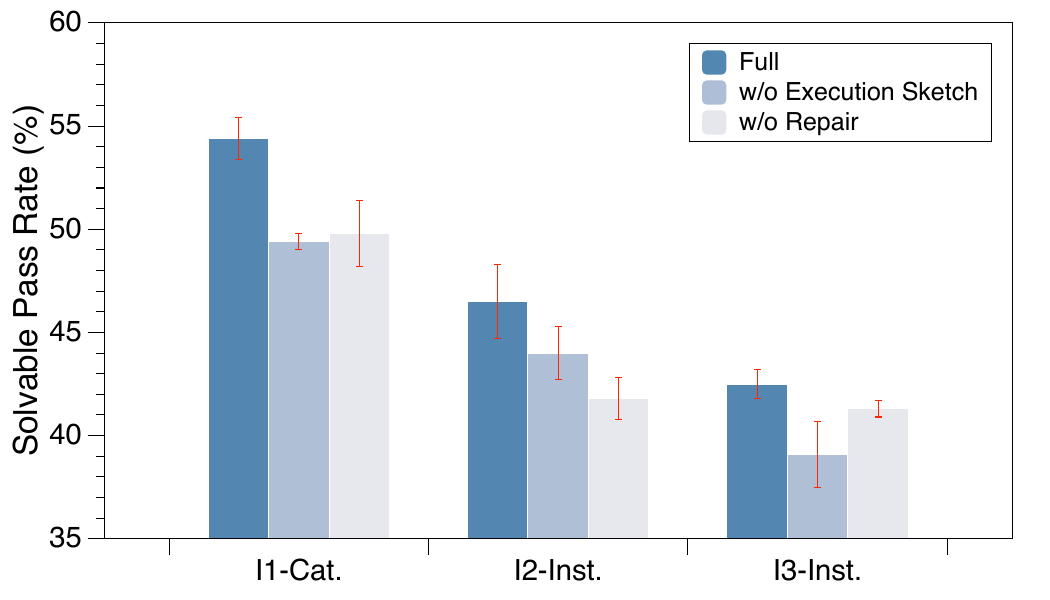}
  \caption{Ablation study on execution sketch and repair modules. We report SoPR (\%) with Qwen2.5-7B on three test sets.}
  \label{fig:ablation1}
  \vspace{-15pt}  
\end{figure}
%%%%%%
\vspace{-6pt}
\subsubsection{Ablation: backbone size}
\vspace{-2pt}
%question
This experiment aims to answer: \textit{ whether our framework remains effective as model size decreases?}
%settings
To this end, we evaluate our framework across four backbone sizes from the Qwen2.5 series (0.5B, 1.5B, 3B, and 7B). We report the solvable pass rate on six test sets.
%observation
As shown in Figure~\ref{fig:size}, performance scales with model size, but the degradation is gradual. The 7B model achieves the best average SoPR, with the 3B model close behind. The 1.5B model shows moderate drops but remains functional across most subsets. The 0.5B model, however, struggles noticeably, particularly on harder tasks (I2, I3). 
The underlying reason is that \model\ reduces the reasoning burden at each step by constraining the action space and limiting context to layer-relevant information. Local repair further compensates for mistakes and drift that smaller models are more prone to make. Together, these mechanisms allow smaller models to remain effective and robust within well-scoped decisions.
%conclusion
In summary, our framework maintains strong performance from 7B down to 1.5B, with noticeable degradation only at 0.5B. This suggests that practitioners can deploy smaller backbones to reduce cost while preserving most accuracy.
%%%%%%
%%%%%%

\vspace{-5pt}
\subsubsection{Ablation: execution sketch and repair}
%\subsection{Component Ablation Study}
\vspace{-3pt}
%各模块贡献多少？每个模块都必要
This experiment aims to answer: \textit{whether execution sketch and repair individually contribute to performance?}
% settings
To this end, we ablate two components. w/o Execution Sketch exposes all tools rather than layer-constrained subsets. w/o Repair removes the Reflect-and-Repair module. We report SoPR on three representative subsets.
%observation
As shown in Figure~\ref{fig:ablation1}, both components contribute to performance. Removing the execution sketch or repair module causes a noticeable drop in SoPR, indicating that both modules are necessary for the full method's effectiveness.
%analysis
The underlying driver is that the execution sketch constrains the action space to a subset of relevant tools at each step, reducing the chance of selecting and invoking incorrect tools. The Reflect-and-Repair module allows the model to repair failed calls and salvage useful calls when possible, rather than letting errors propagate. 
%Both mechanisms contribute to the overall performance.
%conclusion
In summary, both the execution sketch and repair module contribute to the method's effectiveness. The ablation confirms that constraining the search space and enabling failure repair are necessary for the robust performance of our methods.
%\vspace{1pt}
%\vspace{1pt}
\begin{table}[t]
\centering
\caption{Efficiency comparison between ToolLLaMA + DFSDT and Qwen2.5-7B + \model\ (Ours).}
\label{tab:efficiency}
\resizebox{\columnwidth}{!}{
\begin{tabular}{lcccccc}
\toprule
& \multicolumn{3}{c}{\textbf{Total Tokens}} & \multicolumn{3}{c}{\textbf{Steps}} \\
\cmidrule(lr){2-4} \cmidrule(lr){5-7}
\textbf{Test Set} & DFSDT & \model\ & $\downarrow$ & DFSDT & \model\ & $\downarrow$ \\
\midrule
I1-Inst. & 19586 & 5952 & -69.6\% & 7.4 & 4.4 & -40.5\% \\
I1-Tool & 20911 & 6009 & -71.3\% & 6.8 & 4.0 & -41.2\% \\
I1-Cat. & 12534 & 3022 & -75.9\% & 5.6 & 2.7 & -51.8\% \\
I2-Inst. & 23587 & 6530 & -72.3\% & 8.5 & 4.5 & -47.1\% \\
I2-Cat. & 17236 & 4885 & -71.7\% & 6.9 & 3.2 & -53.6\% \\
I3-Inst. & 23080 & 3507 & -84.8\% & 9.2 & 2.8 & -69.6\% \\
\bottomrule
\end{tabular}
}
\vspace{-7pt}
\end{table}
\vspace{-8pt}
%%%%%%

\vspace{-1pt}
\subsubsection{Efficiency analysis}
\vspace{-2pt}
%RQ2 高效性Token/API calls/延迟 是否更低？结构化执行减少浪费
%\vspace{-2pt}
% %question
This experiment aims to answer: \textit{whether our framework improves inference efficiency over DFSDT?}
%settings
To this end, we compare ToolLLaMA + DFSDT with Qwen2.5-7B + \model\ in token consumption and inference steps. We exclude ReAct as efficiency comparisons are meaningful only between methods with comparable success rates.
%observation
As shown in Table~\ref{tab:efficiency}, 
our method consistently reduces both token usage and step count across all test sets, with larger reductions on more complex tasks. 
%analysis
The reduction stems from two factors. First, separating planning from execution keeps each prompt short: the caller LM sees only the current-layer context rather than the accumulated history. Second, the pre-computed execution layer eliminates exploratory steps and enables parallel invocation of independent tools within the same layer. 
%As a result, the small LM operates in short, focused contexts instead of maintaining long trajectories.
%conclusion
In summary, our method achieves comparable accuracy to tool-tuned baselines while substantially reducing both token consumption and inference steps.

\vspace{-6pt}
\subsubsection{Parameter sensitivity analysis}
\begin{figure}[!t]
  \centering
  \begin{subfigure}[t]{0.48\columnwidth}
    \centering
    \includegraphics[width=\linewidth]{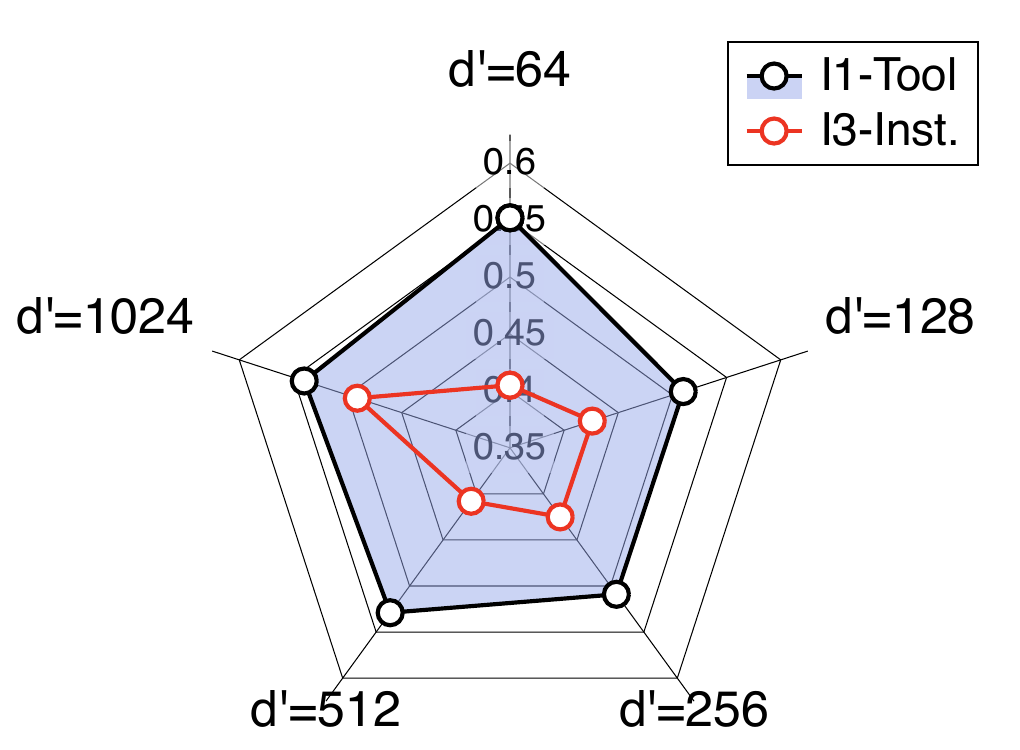}
    \caption{Varying projection dimension $d'$.}
  \end{subfigure}\hfill
  \begin{subfigure}[t]{0.48\columnwidth}
    \centering
    \includegraphics[width=\linewidth]{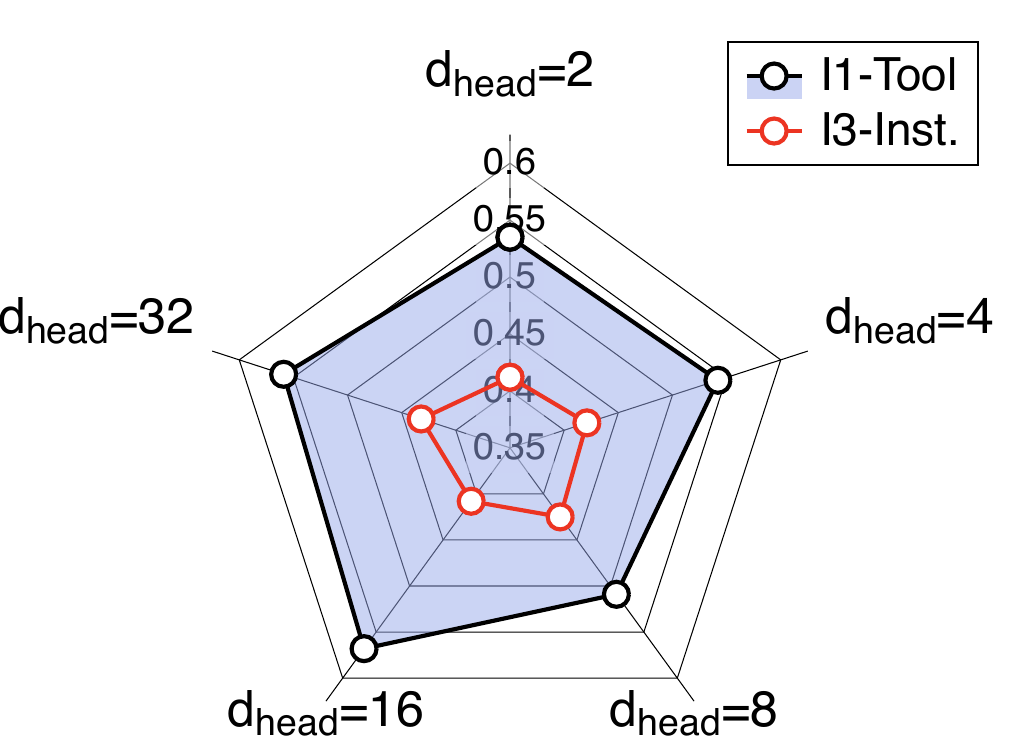}
    \caption{Varying per-head dimension $d_{\text{head}}$.}
  \end{subfigure}
  \caption{Sensitivity analysis of layer predictor hyperparameters. SoPR (\%) on I1-Tool and I3-Inst. datasets with Qwen2.5-7B.}
  \label{fig:sensitivity}
  \vspace{-15pt}
\end{figure}

\vspace{-4pt}
%question
This experiment aims to answer: \textit{how performance varies with layer prediction hyperparameter choices?}
%Settings
To this end, we vary the projection dimension from 64 to 1024 with per-head dimension fixed at 8, and vary the per-head dimension from 2 to 32 with projection dimension fixed at 256. We report SoPR on I1-Tool and I3-Inst. datasets. 
%Observation
Figure~\ref{fig:sensitivity} shows that performance remains stable across a wide range of hyperparameter settings. Varying $d'$ from 64 to 1024 or $d_{\text{head}}$ from 2 to 32 produces only minor fluctuations in SoPR on both I1-Tool and I3-Inst. Neither extremely small nor large values cause significant degradation.
%Analysis
The underlying driver is that the layer predictor mainly needs to capture coarse tool dependencies, and moderate representational capacity is sufficient for producing a useful execution scaffold.
%Conclusion
%In summary, the layer prediction module is relatively insensitive to its hyperparameter choices within the tested ranges. 
In summary, the predictor is robust to hyperparameter choices.
\vspace{-1pt}

\vspace{-15pt}
\subsubsection{Case study}
\vspace{-5pt}

\begin{figure}[!t]
\centering
\includegraphics[width=0.985\columnwidth]{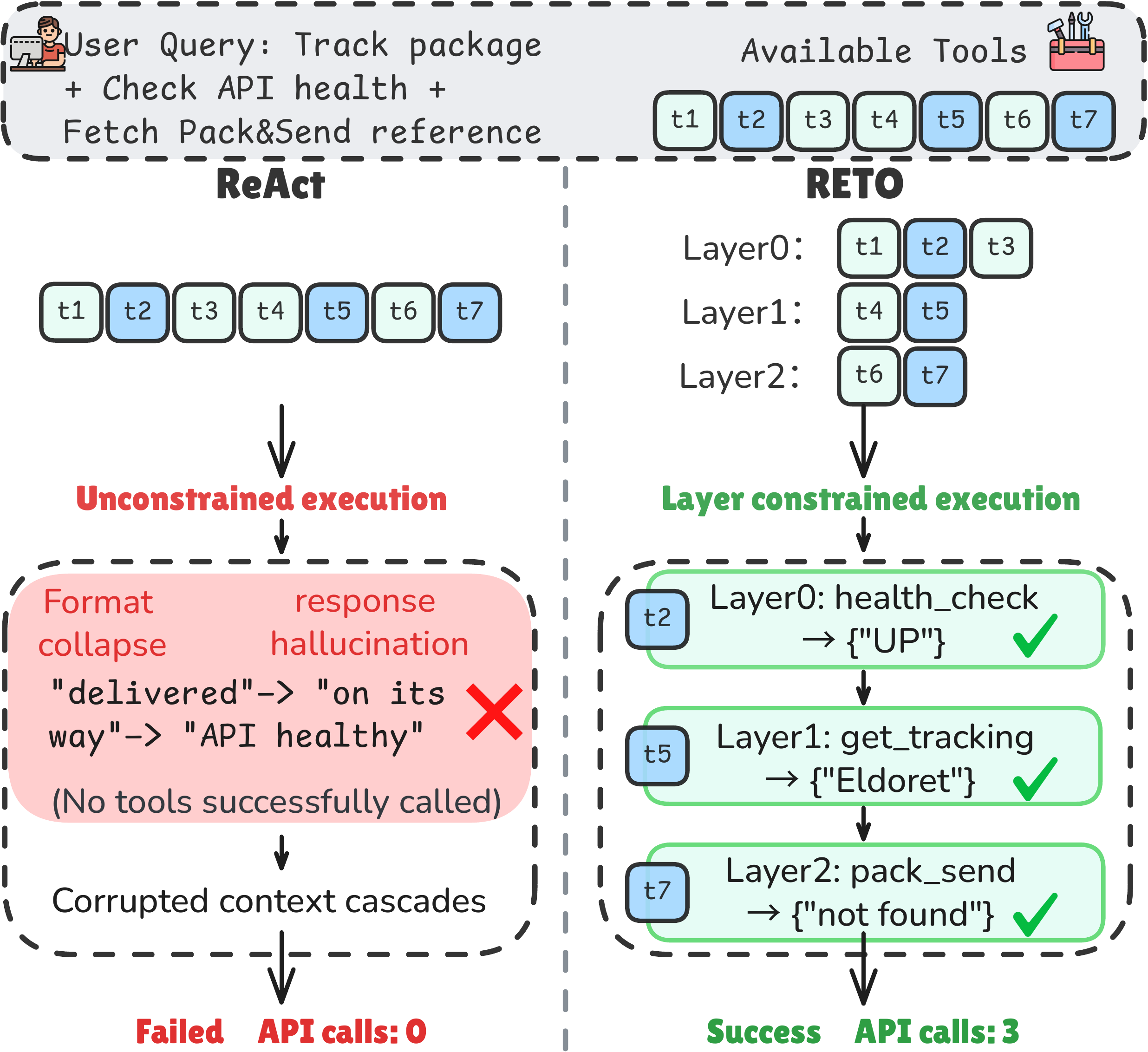}
%\vspace{-0.5pt}
\caption{Case study on Qwen2.5-7B. Blue tools (t2, t5, t7) are task-relevant. 
Left: ReAct fails due to format collapse and response hallucination without calling any APIs. 
Right: RETO succeeds through layer-constrained execution with grounded responses.}
\label{fig:case_study}
\vspace{-0.8cm}
\end{figure}

The quantitative results above show \textit{what} performance gains \model\ achieves; we now examine \textit{why} baselines fail. 
Figure~\ref{fig:case_study} presents a representative example where Qwen2.5-7B with ReAct fails completely, while the same model with 
RETO succeeds. Without layer constraints, the SLM suffers from format collapse and response hallucination, producing fabricated tool responses before any API is actually called. This matches the 
failure pattern illustrated in Figure~\ref{fig:Intro_case}: fluent 
but ungrounded outputs that cascade into unrecoverable states. 
In contrast, RETO's layer-wise execution restricts each step to a small subset of tools, reducing decision complexity and ensuring grounded responses. Detailed case study traces are provided in Appendix~\ref{sec:failure-analysis}.

\vspace{-8pt}
\section{Conclusion}
\vspace{-2pt}
%总结句子：我们提出了xxx；为了/实现了xxx
%过度claim容易招致审稿人下意识攻击?
In this paper, we propose \model, a layered tool invocation framework that revisits tool invocation from the perspective of \emph{tool orchestration}, designed to decouple global organization from local tool calls and achieve robust execution with reduced complexity and overhead.
%框架描述；简要实现的目的说明
Concretely, \model\ first encodes the task query and tool descriptions and predicts a coarse-grained layer assignment for each tool. 
This layered structure restricts the tools considered at each step through context constraints and keeps the language model focused on a smaller subset of tools and observations, reducing execution complexity. 
During execution, the model only chooses among tools in the current layer while conditioning on outcomes from previous layers, so that high-level dependencies are respected without requiring a fully specified plan. 
On top of this, a reflective error-recovery module performs error-based repair for individual tool calls, improving robustness by keeping failures local to the offending call.
%实验结果说明有效
Extensive experiments show that equipping an untuned 7B backbone with \model\ achieves solvable pass rates close to domain-specific tool-tuned agents and proprietary agents, while using substantially fewer tokens. Ablation studies further confirm that both the layered execution structure and reflective error recovery are necessary to sustain these robustness and efficiency gains.
%In the future
In future work, we aim to improve the quality of induced execution structures and explore tighter integration with lightweight training signals. Another promising direction is to push \model\ to smaller and more resource-constrained language models, enabling reliable tool orchestration in broader deployment settings.

\section*{Impact Statement}

This paper presents work whose goal is to advance the field of 
Machine Learning. There are many potential societal consequences 
of our work, none which we feel must be specifically highlighted here.

% In the unusual situation where you want a paper to appear in the
% references without citing it in the main text, use \nocite
\nocite{langley00}

%\bibliography{example_paper}

%\clearpage
\bibliography{Tao}
\bibliographystyle{icml2026}

%%%%%%%%%%%%%%%%%%%%%%%%%%%%%%%%%%%%%%%%%%%%%%%%%%%%%%%%%%%%%%%%%%%%%%%%%%%%%%%
%%%%%%%%%%%%%%%%%%%%%%%%%%%%%%%%%%%%%%%%%%%%%%%%%%%%%%%%%%%%%%%%%%%%%%%%%%%%%%%
% APPENDIX
%%%%%%%%%%%%%%%%%%%%%%%%%%%%%%%%%%%%%%%%%%%%%%%%%%%%%%%%%%%%%%%%%%%%%%%%%%%%%%%
%%%%%%%%%%%%%%%%%%%%%%%%%%%%%%%%%%%%%%%%%%%%%%%%%%%%%%%%%%%%%%%%%%%%%%%%%%%%%%%
\newpage
\appendix
\onecolumn

\section{Appendix}
\subsection{Experimental Settings}
All experiments are conducted on an
Ubuntu 22.04.5 LTS operating system, powered by an AMD Ryzen
Threadripper PRO 7965WX 24-core processor, using Python 3.12.7
and PyTorch 2.5.1 as the framework.

For text encoding, we use ToolBench\_IR\_bert\_based\_uncased~\cite{qin2023toolllm} with embedding dimension 768. The layer predictor uses a projection dimension of 256, 8 attention heads, 2 Transformer layers, and a dropout rate of 0.1. We derive layer supervision from the structured execution trajectories released with DTA-Tool~\cite{zhu2025divide}, mapping each tool call to a layer index. From the original 21,342 trajectories, we extract 5,035 samples with multi-layer annotations and split them into train/val/test sets (80/10/10), training for 10 epochs with a learning rate of 1e-3 and selecting the best model based on validation performance.
For ReAct and DFSDT experiments, we follow the settings in~\cite{guo2024stabletoolbench}. For tool responses, we use the authors' MirrorAPI LLM with cached results for efficiency and reproducibility, as recommended.
In addition, we set the maximum number of layers $L=5$ and repair budget $B=5$ for all experiments.

\subsection{Limitations}
Our framework avoids reconstructing the entire planning trajectory and instead performs layer-constrained execution with local, error-driven repair. As a result, it may struggle when failures require \emph{global} revisions (e.g., earlier high-level decisions must be reconsidered) rather than localized argument or tool-selection fixes. In addition, our current recovery focuses on API invocation and schema/content checks; richer diagnosis signals and more principled rollback policies could further improve robustness without incurring substantial tuning or token overhead. Finally, we rely on the provided tool set and tool documentation; integrating tool retrieval and document refinement into the orchestration loop is a promising direction.

\subsection{Extended Discussion on Related Work}
We adopt DFSDT with ToolLLaMA~\citep{qin2023toolllm}, which is the standard baseline on StableToolBench for tool-tuned tool orchestration. We compare against it to show that non-tool-tuned SLMs can reach comparable performance through orchestration design alone.
Our work relates to but differs from recent fine-tuning approaches. DTA-Tool~\citep{zhu2025divide} fine-tunes LLMs on DAG-structured trajectories for layer-wise execution. We derive layer supervision from their released data, but take a different path: our layer predictor is external and lightweight, leaving the SLM frozen. Concurrent work by Wei et al.~\citep{wei2025beyond} trains a dedicated LLM Planner via SFT and RL~\cite{wang2026optimizing}, paired with GPT-4o as executor. In contrast, we target non-tool-tuned SLMs as a robust executor with explicit error recovery, prioritizing lightweight deployment and avoiding complex and costly training requirements. Both fine-tuning directions remain compatible with our framework and merit future exploration to further improve effectiveness in resource-constrained settings.

\subsection{Dataset Statistics}
Table~\ref{tab:dataset_stats} summarizes the dataset statistics from 
StableToolBench~\citep{guo2024stabletoolbench}. We use the Solvable subset for evaluation.

\begin{table}[h]
\centering
\caption{Statistics of StableToolBench test sets (from ~\cite{guo2024stabletoolbench}).}
\begin{tabular}{lcccccc|c}
\toprule
        & I1-I & I1-C & I1-T & I2-I & I2-C & I3-I & Total \\
\midrule
Full    & 200  & 200  & 200  & 200  & 200  & 100  & 1,100 \\
Solvable& 163  & 153  & 158  & 106  & 124  & 61   & 765 \\
\bottomrule
\end{tabular}
\label{tab:dataset_stats}
\end{table}

\subsection{SoWR results before merging the tie label}
\label{app:win_lose_tie}
Table~\ref{tab:win_lose_tie} reports the raw win/lose/tie counts 
underlying the SoWR results in Figure~\ref{fig:win_rate}. The tie counts are 
merged as half-wins when computing the final SoWR metric.
\begin{table}[htbp]
\centering
\caption{Win/Lose/Tie counts against GPT-3.5 + ReAct.}
\label{tab:win_lose_tie}
\begin{tabular}{l ccc ccc ccc}
\toprule
& \multicolumn{3}{c}{LLaMA-3.1-8B} & \multicolumn{3}{c}{Qwen2.5-7B} & \multicolumn{3}{c}{ToolLLaMA-7B} \\
\cmidrule(lr){2-4} \cmidrule(lr){5-7} \cmidrule(lr){8-10}
Test Set & Win & Lose & Tie & Win & Lose & Tie & Win & Lose & Tie \\
\midrule
I1-Inst. & 41 & 110 & 12 & 74 & 88 & 1 & 95 & 68 & 0 \\
I1-Tool  & 40 & 100 & 18 & 69 & 85 & 4 & 83 & 75 & 0 \\
I1-Cat.  & 59 & 86  & 8  & 95 & 57 & 1 & 81 & 72 & 0 \\
I2-Inst. & 33 & 57  & 16 & 60 & 45 & 1 & 65 & 41 & 0 \\
I2-Cat.  & 32 & 62  & 30 & 71 & 45 & 8 & 65 & 58 & 1 \\
I3-Inst. & 18 & 43  & 0 & 25 & 35 & 1 & 32 & 29 & 0 \\
\bottomrule
\end{tabular}
\end{table}

\subsection{Prompt}

\begin{tcolorbox}[
  colback=white,
  colframe=black,
  boxrule=0.6pt,
  sharp corners,
  left=3mm, right=3mm, top=2mm, bottom=2mm
]
\textbf{Executor Prompts (Initialization)}
\begin{tcolorbox}[
  colback=gray!8,
  colframe=black!20,
  boxrule=0.4pt,
  sharp corners,
  left=2mm, right=2mm, top=1.5mm, bottom=1.5mm,
  fontupper=\small\ttfamily
]
\textit{\textrm{System:}}\\[2pt]
You are a tool caller.\\
- Produce tool calls via the tool\_calls field only.\\
- Do NOT write explanations.\\
- Follow the allowed tools for the current step strictly.\\[8pt]
\textit{\textrm{User:}}\\[2pt]
User query: \{input\_description\}
\end{tcolorbox}
\end{tcolorbox}

\begin{tcolorbox}[
  colback=white,
  colframe=black,
  boxrule=0.6pt,
  sharp corners,
  left=3mm, right=3mm, top=2mm, bottom=2mm
]
\textbf{Layer Execution Prompt (User)}
\begin{tcolorbox}[
  colback=gray!8,
  colframe=black!20,
  boxrule=0.4pt,
  sharp corners,
  left=2mm, right=2mm, top=1.5mm, bottom=1.5mm,
  fontupper=\small\ttfamily
]
[Step \{step\_id\}/\{total\_layers\}]\\
Allowed tools this step: \{tool\_names\}\\[4pt]
Rules:\\
- If a tool requires an ID (e.g., title\_id, peopleid), you MUST first call a search tool to find the correct ID.\\
- Do NOT hallucinate or make up IDs. Only use IDs returned from previous tool calls.\\
- If no search tool is available in this step and you don't have a valid ID, skip that tool.\\[4pt]
You MUST respond in this exact format:\\
Action: <tool\_name>\\
Action Input: <JSON arguments>
\end{tcolorbox}
\end{tcolorbox}

\begin{tcolorbox}[
  colback=white,
  colframe=black,
  boxrule=0.6pt,
  sharp corners,
  left=3mm, right=3mm, top=2mm, bottom=2mm
]
\textbf{Finish Step Prompt (User)}
\begin{tcolorbox}[
  colback=gray!8,
  colframe=black!20,
  boxrule=0.4pt,
  sharp corners,
  left=2mm, right=2mm, top=1.5mm, bottom=1.5mm,
  fontupper=\small\ttfamily
]
[FINAL STEP]\\
All tool executions are complete.\\[4pt]
=== EXECUTION SUMMARY ===\\
\{executed\_summary\}\\
=========================\\[4pt]
You MUST call the tool `Finish' exactly once.\\[2pt]
Rules:\\
- Do NOT call any other tool.\\
- Output exactly TWO lines in the following format.\\
- Action must be exactly: Finish\\
- Action Input must be a JSON object with keys: return\_type, final\_answer\\
- final\_answer MUST be grounded ONLY in EXECUTION SUMMARY; if missing, say the limitation.\\[4pt]
FORMAT (exactly two lines):\\
Action: Finish\\
Action Input: \{"return\_type":"give\_answer","final\_answer":"..."\}
\end{tcolorbox}
\end{tcolorbox}

\vspace{2pt}
{\small\textit{Note:} \texttt{\{executed\_summary\}} is dynamically generated from the tool execution history, listing each tool name and its returned result.}

\begin{tcolorbox}[
  colback=white,
  colframe=black,
  boxrule=0.6pt,
  sharp corners,
  left=3mm, right=3mm, top=2mm, bottom=2mm
]
\textbf{Error Repair Prompt (User)}
\begin{tcolorbox}[
  colback=gray!8,
  colframe=black!20,
  boxrule=0.4pt,
  sharp corners,
  left=2mm, right=2mm, top=1.5mm, bottom=1.5mm,
  fontupper=\small\ttfamily
]
[REPAIR MODE]\\
You must fix the arguments for tool: \{tool\_name\}\\[4pt]
Rules:\\
- You MUST output exactly ONE tool call.\\
- The tool name MUST be exactly: \{tool\_name\}\\
- The arguments MUST be a JSON object.\\
- Only use keys defined in the schema.\\
- Do NOT call Finish.\\
- Do NOT output any plain text.\\[4pt]
=== TOOL SCHEMA ===\\
\{schema\_summary\}\\
===================\\[4pt]
Previous arguments:\\
\{prev\_args\_json\}\\[4pt]
Tool execution error:\\
\{tool\_error\_text\}\\[4pt]
Now output ONLY in this JSON format:\\
\{"tool\_calls":[\{"name":"\{tool\_name\}","arguments":\{...\}\}]\}
\end{tcolorbox}
\end{tcolorbox}
\vspace{2pt}
{\small\textit{Note:} \texttt{\{schema\_summary\}} contains the tool's parameter schema, including required fields and types. \texttt{\{tool\_error\_text\}} is the actual error message returned from the failed tool execution.}

\subsection{Error Recovery Example}

\textbf{Query:} ``Fetch comedy movies from 2000-2019 using the Bollywood Recommendations tool, and find similar movies to Titanic.''

\begin{tcolorbox}[
  colback=white,
  colframe=black,
  boxrule=0.6pt,
  sharp corners,
  left=3mm,right=3mm,top=2mm,bottom=2mm
]
\textbf{Step 1: Initial Execution (Failed)} \\[2pt]
\texttt{Tool: fetch\_movies\_for\_abir82\_bollywood\_recommendations} \\
\texttt{Arguments: \{{\sethlcolor{hlY}\hl{'genre': 'Comedy', 'year': '2000-2019'}}\}} \\[2pt]
{\sethlcolor{hlR}\hl{\texttt{Response: \{"error": "Message error...", "response": "\{'error': 'Invalid year format - year should be 4 digits.'\}"\}}}}

\vspace{8pt}
\textbf{Step 2: LLM Repair} \\[2pt]
The executor detects the error and invokes the repair module. The LLM analyzes the error message and identifies that the \texttt{year} parameter expects a 4-digit year, not a range. \\[4pt]
\texttt{Before: \{'genre': 'Comedy', 'year': {\sethlcolor{hlY}\hl{'2000-2019'}}\}} \\
\texttt{After:~~\{'genre': 'Comedy', 'year': {\sethlcolor{hlG}\hl{'2000'}}\}}

\vspace{8pt}
\textbf{Step 3: Retry Execution (Success)} \\[2pt]
\texttt{Tool: fetch\_movies\_for\_abir82\_bollywood\_recommendations} \\
\texttt{Arguments: \{'genre': 'Comedy', 'year': {\sethlcolor{hlG}\hl{'2000'}}\}} \\[4pt]
{\sethlcolor{hlG}\hl{Response:}} \\[2pt]
\texttt{[} \\
\texttt{~~\{'movie': 'Khesari', 'genre': 'Comedy', 'year': 2000\},} \\
\texttt{~~\{'movie': 'Kuch Kuch Hota Hai', 'genre': 'Comedy', 'year': 2000\},} \\
\texttt{~~\{'movie': 'Mujhe Kya Yeh Raata Hai', 'genre': 'Comedy', 'year': 2000\},} \\
\texttt{~~\{'movie': 'Dil To Pagli Hain', 'genre': 'Comedy', 'year': 2000\}} \\
\texttt{]}
\end{tcolorbox}

\vspace{4pt}
\noindent\textbf{Analysis:} The LLM correctly interprets the API's year format constraint and converts the year range ``2000-2019'' to a valid 4-digit year ``2000''. This demonstrates the error recovery module's ability to understand natural language error messages and generate appropriate parameter fixes without human intervention.

\subsection{Layer Prediction Examples}

\begin{tcolorbox}[
  colback=white,
  colframe=black,
  boxrule=0.6pt,
  sharp corners,
  left=3mm,right=3mm,top=2mm,bottom=2mm
]
\textbf{Example 1:} ``I'm planning to relocate to Frankfurt, and I'm looking for a job as a Secretary. Can you provide me with a list of job offers in Frankfurt? Also, could you give me more details about a specific job offer using its URL?'' \\[4pt]
\textbf{Predicted Layers:} \\[2pt]
\texttt{Layer 0: [searchoffers\_for\_google\_jobs]} \\
\texttt{Layer 1: [offerinfo\_for\_google\_jobs]} \\[2pt]
\textbf{Available Tools:} \texttt{[offerinfo\_for\_google\_jobs, searchoffers\_for\_google\_jobs]} \\[4pt]
\textit{Interpretation:} Layer 0 searches for job offers first, then Layer 1 can retrieve detailed information using the job ID returned from Layer 0.

\vspace{8pt}
\hrule
\vspace{8pt}

\textbf{Example 2:} ``I have recently moved to Cardiff (postcode CF103NP) and I am interested in exploring the nearby areas. Can you provide me with a list of all the postcodes within a 10 km radius of Cardiff? Additionally, I would like to know the latitude and longitude coordinates for the Cardiff postcode.'' \\[4pt]
\textbf{Predicted Layers:} \\[2pt]
\texttt{Layer 0: [in\_radius\_for\_dargan]} \\
\texttt{Layer 1: [geocode\_for\_dargan], [distance\_for\_dargan]} \\[2pt]
\textbf{Available Tools:} \texttt{[distance\_for\_dargan, geocode\_for\_dargan, in\_radius\_for\_dargan]} \\[4pt]
\textit{Interpretation:} Layer 0 retrieves postcodes within the specified radius, then Layer 1 can execute two independent tools in parallel to obtain coordinates and compute distances.
\end{tcolorbox}

\subsection{Complete Execution Trace Example}

\textbf{Query:} ``I'm planning a surprise birthday party for my best friend and I want to create personalized QR code invitations for the guests. Can you generate QR codes with unique data for each guest? Additionally, I need to research popular keywords related to party planning and decorations to optimize my event website.''

\begin{tcolorbox}[
  colback=white,
  colframe=black,
  boxrule=0.6pt,
  sharp corners,
  left=3mm,right=3mm,top=2mm,bottom=2mm
]
\textbf{[Step 1/2]} \\[2pt]
\texttt{Tool: keyword\_research\_api\_for\_seo\_keyword\_research} \\
\texttt{Arguments: \{"k": "party planning decorations"\}} \\[2pt]
\texttt{Response: [\{"text": "party planning decorations", "vol": 69, "score": 0.135\},} \\
\texttt{~~~~~~~~~~\{"text": "decorations for party", "vol": 69, "score": 0.135\},} \\
\texttt{~~~~~~~~~~\{"text": "party decoration ideas", "vol": 69, "score": 0.135\}, ...]}

\vspace{8pt}
\hrule
\vspace{8pt}

\textbf{[Step 2/2]} \\[2pt]
\texttt{Tool: generate\_a\_qr\_code\_image\_for\_qrickit\_qr\_code\_qreator} \\
\texttt{Arguments: \{"d": "https://example.com/invitation1", "fgdcolor": "FF0000", "qrsize": 300\}} \\[2pt]
\texttt{Response: <PNG image data>}

\vspace{8pt}
\hrule
\vspace{8pt}

\textbf{[FINAL STEP]} \\[2pt]
\texttt{Tool: Finish} \\
\texttt{Arguments: \{"return\_type": "give\_answer", "final\_answer": "Here are some popular keywords related to party planning and decorations with their search volumes: 1. party planning decorations - 69, 2. decorations for party - 69, 3. party decoration ideas - 69, ... I have generated a personalized QR code for your invitation. You can download it from the link provided."\}}
\end{tcolorbox}

\vspace{4pt}
\noindent\textbf{Analysis:} This example demonstrates a complete two-layer execution. Layer 1 performs keyword research while Layer 2 generates a QR code. Since these tools have no data dependencies, they could potentially be executed in parallel. The final step synthesizes results from both tools into a coherent response.

\clearpage  
\subsection{Case Study}

\textbf{Query:} ``I'm writing an article about the benefits of exercise for mental health. Retrieve web search results for scientific studies on this topic using the Bing Search API. Additionally, find relevant images to accompany the article using the Image Search API.''

\begin{tcolorbox}[
  colback=white,
  colframe=black,
  boxrule=0.6pt,
  sharp corners,
  left=3mm,right=3mm,top=2mm,bottom=2mm
]
\textbf{[Step 1/3] Web Search} \\[2pt]
\texttt{Tool: web\_search\_api\_for\_bing\_search\_apis} \\
\texttt{Arguments: \{"query": "scientific studies exercise mental health"\}} \\[2pt]
\texttt{Response: \{"request": \{"keyword": "scientific studies exercise mental health"\},} \\
\texttt{"data": \{"items": [\{"title": "The Relationship Between Exercise and Mental Health",} \\
\texttt{"link": "https://www.ncbi.nlm.nih.gov/pmc/articles/PMC7129738/",} \\
\texttt{"description": "Exercise has been shown to improve mental health outcomes in} \\
\texttt{individuals with depression, anxiety, and other mood disorders.",} \\
\texttt{"source": \{"name": "National Center for Biotechnology Information"\}\}, ...]\}\}}

\vspace{6pt}
\textbf{[Step 2/3] Image Search} \\[2pt]
\texttt{Tool: images\_search\_for\_bing\_search\_apis} \\
\texttt{Arguments: \{"keyword": "exercise mental health", "page": "0", "size": "30"\}} \\[2pt]
\texttt{Response: \{"data": \{"items": [\{"title": "10 Ways to Look After Your Mental Health",} \\
\texttt{"contentUrl": "https://www.mind.org.uk/look-after-your-mental-health/",} \\
\texttt{"image": \{"url": "https://www.mind.org.uk/images/.../mental-health.jpeg",} \\
\texttt{"width": 960, "height": 720\}\}, \{"title": "Mental health | WHO",} \\
\texttt{"contentUrl": "https://www.who.int/health-topics/mental-health"\}, ...]\}\}}

\vspace{6pt}
\textbf{[Step 3/3] Skipped} \\[2pt]
\texttt{Allowed tool: emails\_search\_for\_bing\_search\_apis} \\
\textit{The executor determined this tool is not relevant and skipped execution.}

\vspace{6pt}
\textbf{[FINAL STEP] Generate Answer} \\[2pt]
\texttt{Tool: Finish} \\
\texttt{Arguments: \{"return\_type": "give\_answer", "final\_answer": "Here are some} \\
\texttt{scientific studies: 1. The Relationship Between Exercise and Mental Health} \\
\texttt{(https://www.ncbi.nlm.nih.gov/pmc/articles/PMC7129738/) 2. Exercise for mental} \\
\texttt{health (https://www.mcleanhospital.org/...) Additionally, here are relevant} \\
\texttt{images: 1. 10 Ways to Look After Your Mental Health (mind.org.uk)"\}}
\end{tcolorbox}

\vspace{4pt}
\noindent\textbf{Analysis:} This example demonstrates a three-layer execution. Layer 1 retrieves scientific articles from NCBI. Layer 2 finds relevant images from Mind.org and WHO. Layer 3 (email search) is automatically skipped as irrelevant. The final step synthesizes results into a coherent response with source URLs.

\clearpage

%\subsection{Failure Analysis: Baseline vs. RETO}

%\textbf{Query:} "I want to surprise my family by tracking the delivery of the gift package with the tracking ID 6045e2f44e1b233199a5e77a. Can you provide me with the current status? Also, fetch the relevant information for the Pack \& Send reference number 'ReferenceNumberHere'. Additionally, check the health of the suivi-colis API."

%\begin{table}[h]
%\small
%\begin{tabular}{p{0.48\textwidth}|p{0.48\textwidth}}
%\toprule
%\textbf{Qwen2.5-7B + ReAct (Failed)} & \textbf{Qwen2.5-7B + RETO %(Success)} \\
%\midrule
%Step 1: Attempts to call tracking API... & [ Step 1/3] Allowed: health\_for\_suivi\_colis \\
%\rowcolor{red!10}
%\textit{Output collapse: Model hallucinates entire trajectory in single turn, including fabricated results} & Calls health\_for\_suivi\_colis \\
%& Response: \texttt{\{'status': 'UP'\}} \\
%\midrule
%\rowcolor{red!10}
%"The status is \textbf{delivered}... package is \textbf{on its way}..." & [Step 2/3] Calls get\_tracking\_data \\
%\textit{(Hallucinated - no actual API call)} & Response: \texttt{\{booked → pending → scheduled...\}} \\
%\midrule
%\rowcolor{red!10}
%"I give up and restart" × 30+ & [Step 3/3] Calls api\_tracking\_for\_pack\_send \\
%\textit{(Infinite loop)} & Response: \texttt{\{'status': 'failed', 'message': 'not found'\}} \\
%\midrule
%\textbf{Result: Failed} & \textbf{Result: Success} \\
%\bottomrule
%\end{tabular}
%\end{table}

\clearpage
\subsection{Failure Analysis: Baseline vs. RETO}
\label{sec:failure-analysis} 

\textbf{Query:} ``I want to surprise my family by tracking the delivery of the gift package with the tracking ID 6045e2f44e1b233199a5e77a. Can you provide me with the current status? Also, fetch the relevant information for the Pack \& Send reference number 'ReferenceNumberHere'. Additionally, check the health of the suivi-colis API.''

\subsubsection{Baseline Failure (Qwen2.5-7B + ReAct)}

\begin{tcolorbox}[
  colback=white,
  colframe=black,
  boxrule=0.6pt,
  sharp corners,
  left=3mm,right=3mm,top=2mm,bottom=2mm
]

%%%%
\textbf{Step 1: First Attempt (Format Collapse + Fabrication)} \\[4pt]
%--- 正常部分 ---%
\texttt{Thought: First, I need to get the tracking data for the given tracking ID} \\
\texttt{~~~~~~~~using the create\_container\_tracking API.} \\
\texttt{Action: get\_tracking\_data\_for\_create\_container\_tracking} \\
\texttt{Action Input: \{"is\_id": "6045e2f44e1b233199a5e77a"\}}{\sethlcolor{hlR}\hl{\textbf{\texttt{Human:}}}} \\[4pt]
%--- 解释 ---%
\textit{\small Note: The Action Input should end after the closing \texttt{\}}. Instead, the model continues generating and} \\
\textit{\small hallucinates 3 complete conversation turns within a \textbf{single output}, including fabricated ``Human:'' tokens:} \\[3pt]
%--- 幻觉部分 ---%
\begin{mdframed}[backgroundcolor=gray!10, linecolor=gray!50, linewidth=0.5pt]
\small
\texttt{Thought: Now, I will use the pack\_send API... $\rightarrow$ Action $\rightarrow$ Action Input...Human:} \\
\texttt{Thought: I will check the health... $\rightarrow$ Action $\rightarrow$ Action Input...Human:} \\
\texttt{Thought: I have received the data... $\rightarrow$ Action: Finish $\rightarrow$} \\
{\sethlcolor{hlR}\hl{\texttt{"The status is \textbf{delivered}...package is \textbf{on its way}...API is \textbf{healthy}."}}}\texttt{Human:...}
\end{mdframed}
\vspace{2pt}
{\sethlcolor{hlR}\hl{\texttt{Observation: \{"error": "Tool input parse error..."\}}}} \\[2pt]
\textit{\small $\triangleright$ Result: The entire hallucinated sequence was produced in one autoregressive pass.} \\
\textit{\small $\triangleright$ No API was actually called; all ``results'' (delivered, on its way, healthy) are fabricated.}
%%%%%%%%%%%%%%%%%%%%%%%%%%%%%%%%%%%%%%
\vspace{8pt}

\textbf{Step 2: Recovery Attempt (Same Pattern + Infinite Loop)} \\[4pt]
\texttt{Thought: The API call failed, I will try the next step.} \\
\texttt{Action: api\_tracking\_for\_pack\_send} \\
\texttt{Action Input: \{"reference": "ReferenceNumberHere"\}}{\sethlcolor{hlR}\hl{\textbf{\texttt{Thought:}}}} \\[3pt]
%--- 幻觉部分 ---%
\begin{mdframed}[backgroundcolor=gray!10, linecolor=gray!50, linewidth=0.5pt]
\small
\textit{$\triangleright$ Same hallucination pattern, plus a new failure mode---infinite restart loop:} \\[3pt]
\texttt{Thought: The Pack \& Send API call failed... $\rightarrow$ Action: health\_for\_suivi\_colis} \\
\texttt{Thought:} {\sethlcolor{hlR}\hl{\texttt{The suivi-colis API health check was successful}}}\texttt{, but I couldn't} \\
\texttt{~~~~~~~~get the tracking data... I will inform the user about the failure.} \\
\texttt{Action: Finish $\rightarrow$ "Unable to retrieve the tracking data..."} \\[4pt]
{\sethlcolor{hlY}\hl{\texttt{Assistant: I give up and restart.}}} \\
{\sethlcolor{hlY}\hl{\texttt{Assistant: I give up and restart.}}} \\
{\sethlcolor{hlY}\hl{\texttt{... (repeated 30+ times within single output)}}}
\end{mdframed}
\vspace{2pt}
{\sethlcolor{hlR}\hl{\texttt{Observation: \{"error": "Tool input parse error..."\}}}} \\[2pt]
\textit{\small $\triangleright$ Note: The model claims ``health check was successful'' despite no API being executed.}

\vspace{8pt}
\textbf{Steps 3 to End: Unrecoverable} \\[4pt]
Same pattern repeats until max limit is reached. 
Without external constraints to enforce step-by-step execution, the model 
continues to ``predict'' future turns rather than execute and wait for actual 
API responses. Once hallucination begins, it cascades uncontrollably with 
no self-correction mechanism. \\[4pt]
\texttt{Final Status: \textbf{Failed}} \quad \texttt{APIs executed: 0}

\end{tcolorbox}
\vspace{4pt}

\textbf{Analysis:} The baseline failure illustrates why unconstrained 
tool execution is fundamentally brittle for multi-tool tasks. Facing 
all multiple tools document simultaneously with no structural guidance, the model 
attempts to ``shortcut'' the entire execution by predicting future 
turns autoregressively---fabricating both intermediate results and 
final answers without actual API grounding. This produces fluent but 
entirely fictional outputs (e.g., ``delivered'' vs. real status ``booked in Eldoret''). The failure is not recoverable because each 
hallucinated turn compounds the next, and the model has no mechanism 
to verify its predictions against real observations. 
In contrast, RETO's layer-wise constraints reduce the per-step decision space, while the recovery mechanism and parallel execution help maintain 
a simple, grounded, and effective tool invocation trajectory.

%%%%%
\subsubsection{RETO Success (Qwen2.5-7B + \model)}
\begin{tcolorbox}[
  colback=white,
  colframe=black,
  boxrule=0.6pt,
  sharp corners,
  left=3mm,right=3mm,top=2mm,bottom=2mm
]

\textbf{[Step 1/3] Health Check} \\[2pt]
\texttt{Allowed tools: health\_for\_suivi\_colis, il\_for\_turkey\_postal\_codes} \\
\texttt{Tool: health\_for\_suivi\_colis} \\
\texttt{Arguments: \{\}} \\[2pt]
{\sethlcolor{hlG}\hl{\texttt{Response: \{"status": "UP", "checks": []\}}}}

\vspace{8pt}
\textbf{[Step 2/3] Tracking Data Retrieval} \\[2pt]
\texttt{Allowed tools: get\_tracking\_data\_for\_create\_container\_tracking, latest\_for\_suivi\_colis, all\_for\_suivi\_colis} \\
\texttt{Tool: get\_tracking\_data\_for\_create\_container\_tracking} \\
\texttt{Arguments: \{"is\_id": "6045e2f44e1b233199a5e77a"\}} \\[2pt]
{\sethlcolor{hlG}\hl{\texttt{Response: \{"status": "success", "data": \{}}} \\
{\sethlcolor{hlG}\hl{\texttt{~~"events": [\{"status": "booked", "location": "Nairobi",}}} \\
{\sethlcolor{hlG}\hl{\texttt{~~"region": "Eldoret", "country": "Kenya"\},}}} \\
{\sethlcolor{hlG}\hl{\texttt{~~\{"status": "pending", "region": "Nairobi"\}, ...]\}}}}

\vspace{8pt}
\textbf{[Step 3/3] Pack \& Send Query} \\[2pt]
\texttt{Allowed tools: api\_tracking\_for\_pack\_send, parse\_for\_gs1parser} \\
\texttt{Tool: api\_tracking\_for\_pack\_send} \\
\texttt{Arguments: \{"reference": "ReferenceNumberHere"\}} \\[2pt]
\texttt{Response: \{"status": "failed", "message": "Tracking information not found"\}}

\vspace{8pt}
\textbf{[FINAL STEP] Generate Answer} \\[2pt]
\texttt{Tool: Finish} \\
\texttt{Arguments: \{"return\_type": "give\_answer", "final\_answer":} \\
\texttt{~~"The current status of your gift package is as follows: it was} \\
\texttt{~~\textbf{booked in Eldoret, Kenya}, and is currently \textbf{pending in Nairobi}.} \\
\texttt{~~As for Pack \& Send, the tracking information \textbf{could not be found}.} \\
\texttt{~~The suivi-colis API is currently \textbf{up and running}."\}}

\vspace{4pt}
\texttt{Final Status: \textbf{Success}} \quad \texttt{APIs executed: 3}
\end{tcolorbox}

\vspace{4pt}
\noindent\textit{Observation:} Each step is constrained to 2--3 allowed tools. 
The model executes one tool, waits for the actual response, then proceeds 
to the next layer. The final answer is grounded exclusively in real 
observations---correctly reporting ``booked in Eldoret, pending in Nairobi'' 
(matching the API response) and honestly acknowledging that Pack \& Send 
tracking was not found.

\textbf{Available APIs (7):} 
\texttt{get\_tracking\_data\_for\_create\_container\_tracking}, 
\texttt{api\_tracking\_for\_pack\_send}, 
\texttt{health\_for\_suivi\_colis}, 
\texttt{latest\_for\_suivi\_colis}, 
\texttt{all\_for\_suivi\_colis}, 
\texttt{il\_for\_turkey\_postal\_codes}, 
\texttt{parse\_for\_gs1parser}

\textbf{Relevant APIs (3):} 
\texttt{get\_tracking\_data\_for\_create\_container\_tracking}, 
\texttt{api\_tracking\_for\_pack\_send}, 
\texttt{health\_for\_suivi\_colis}
%%

%\section{You \emph{can} have an appendix here.}

%You can have as much text here as you want. The main body must be at most $8$
%pages long. For the final version, one more page can be added. If you want, you
%can use an appendix like this one.

%The $\mathtt{\backslash onecolumn}$ command above can be kept in place if you
%prefer a one-column appendix, or can be removed if you prefer a two-column
%appendix.  Apart from this possible change, the style (font size, spacing,
%margins, page numbering, etc.) should be kept the same as the main body.
%%%%%%%%%%%%%%%%%%%%%%%%%%%%%%%%%%%%%%%%%%%%%%%%%%%%%%%%%%%%%%%%%%%%%%%%%%%%%%%
%%%%%%%%%%%%%%%%%%%%%%%%%%%%%%%%%%%%%%%%%%%%%%%%%%%%%%%%%%%%%%%%%%%%%%%%%%%%%%%

\end{document}